\documentclass[12pt,a4paper,twoside]{article}
\makeatletter\if@twocolumn\PassOptionsToPackage{switch}{lineno}\else\fi\makeatother

\usepackage{amsfonts,amssymb,amsbsy,latexsym,amsmath,tabulary,graphicx,times,xcolor}
\usepackage[T1]{fontenc}
\usepackage{url,multirow,morefloats,floatflt,cancel,tfrupee,multicol}
\usepackage{circledsteps}
\makeatletter

\AtBeginDocument{\@ifpackageloaded{textcomp}{}{\usepackage{textcomp}}}
\makeatother
\usepackage{colortbl}
\usepackage{xcolor}
\usepackage{pifont}
\usepackage[nointegrals]{wasysym}
\usepackage[superscript,biblabel]{cite}
\makeatletter

\usepackage{soul}      
\usepackage{xcolor}    
\sethlcolor{yellow}    

\@ifundefined{etal}{\def\etal{\textit{et~al}}}{}

\AtBeginDocument{
\expandafter\ifx\csname eqalign\endcsname\relax
\def\eqalign#1{\null\vcenter{\def\\{\cr}\openup\jot\m@th
  \ialign{\strut$\displaystyle{##}$\hfil&$\displaystyle{{}##}$\hfil
      \crcr#1\crcr}}\,}
\fi
}

\@ifundefined{tsGraphicsScaleX}{\gdef\tsGraphicsScaleX{1}}{}
\@ifundefined{tsGraphicsScaleY}{\gdef\tsGraphicsScaleY{.9}}{}
\def\checkGraphicsWidth{\ifdim\Gin@nat@width>\linewidth
	\tsGraphicsScaleX\linewidth\else\Gin@nat@width\fi}

\def\checkGraphicsHeight{\ifdim\Gin@nat@height>.9\textheight
	\tsGraphicsScaleY\textheight\else\Gin@nat@height\fi}

\def\fixFloatSize#1{}
\let\ts@includegraphics\includegraphics

\def\inlinegraphic[#1]#2{{\edef\@tempa{#1}\edef\baseline@shift{\ifx\@tempa\@empty0\else#1\fi}\edef\tempZ{\the\numexpr(\numexpr(\baseline@shift*\f@size/100))}\protect\raisebox{\tempZ pt}{\ts@includegraphics{#2}}}}

\AtBeginDocument{\def\includegraphics{\@ifnextchar[{\ts@includegraphics}{\ts@includegraphics[width=\checkGraphicsWidth,height=\checkGraphicsHeight,keepaspectratio]}}}

\DeclareMathAlphabet{\mathpzc}{OT1}{pzc}{m}{it}

\def\URL#1#2{\@ifundefined{href}{#2}{\href{#1}{#2}}}

\edef\fntEncoding{\f@encoding}

\makeatother

\newif\ifmultipleabstract\multipleabstractfalse%
\usepackage[hmargin=1.8cm,vmargin=2.2cm]{geometry}
\makeatletter
\def\author#1{\gdef\@author{\hskip-\dimexpr(\tabcolsep)\hskip1pt\parbox{\dimexpr\textwidth-1pt}{\centering #1}}}

\let\@articletype\@empty \def\articletype#1{\gdef\@articletype{{\fontsize{14}{16}\selectfont #1}}}

\usepackage{fancyhdr}

\fancypagestyle{headings}{\fancyhf{}\fancyhead[RE]{\MakeTextUppercase{\textbf{\RunningAuthor}}}\fancyhead[LE]{\textbf{\thepage}}\fancyhead[LO]{\MakeTextUppercase{\textbf{\RunningHead}}}\fancyhead[RO]{\textbf{\thepage}}}
\fancypagestyle{plain}{\fancyhf{}\fancyfoot[C]{\textbf{\thepage}}}

\AtBeginDocument{\usepackage{lastpage}}
\def\title#1{%
  \gdef\@title{%
    \ifx\@articletype\@empty\else\@articletype~\\\fi%
     #1}%
}

\usepackage{abstract}
\def\abstractname{\textbf{Abstract}}
\renewenvironment{onecolabstract}
{\vspace*{-.4pc}\trivlist\item[]\leftskip1pt\noindent\selectfont\hfill\abstractname\hfill\mbox{\null}\par\ignorespaces}{\endtrivlist}

\def\NormalBaseline{\def\baselinestretch{1.1}}

\usepackage{textcase}
\usepackage[explicit]{titlesec}
\titleformat{\section}[block]{\NormalBaseline\boldmath\bfseries}
{\thesection.}
{6pt}
{#1}
[]
\titleformat{\subsection}[hang]{\NormalBaseline\filright\itshape}
{\thesubsection.}
{6pt}
{#1}
[]
\titleformat{\subsubsection}[runin]{\NormalBaseline\filright\itshape}
{\hspace{16pt}\thesubsubsection}
{6pt}
{#1}
[]
\titleformat{\paragraph}[runin]{\NormalBaseline}
{\theparagraph}
{6pt}
{#1}
[]
\titleformat{\subparagraph}[runin]{\NormalBaseline}
{\thesubparagraph}
{6pt}
{#1}
[]

\titlespacing{\section}{0pt}{1.5\baselineskip}{.2\baselineskip}  
\titlespacing{\subsection}{0pt}{1.5\baselineskip}{.2\baselineskip}  
\titlespacing{\subsubsection}{0pt}{1.5\baselineskip}{.2\baselineskip}  
\titlespacing{\paragraph}{0pt}{.5\baselineskip}{10pt}  
\titlespacing{\subparagraph}{0pt}{.5\baselineskip}{10pt}

\usepackage{caption}
\DeclareCaptionLabelFormat{figlabel}{Figure #2}
\date{}
\makeatother

\usepackage{endfloat}
\usepackage{float}

\usepackage{url} 
\usepackage{booktabs}

\begin{document}
	
\title{Multi-function Robotized Surgical Dissector for Endoscopic Pulmonary Thromboendarterectomy: Preclinical Study and Evaluation}
\def\RunningHead{
	Robotic Dissector for PTE: Preclinical Study and Evaluation
}
\def\RunningAuthor{Zhu \etal}
\author{
	Runfeng Zhu\textsuperscript{1}, 
	Xin Zhong\textsuperscript{2}, 
	Qingxiang Zhao\textsuperscript{1}, 
	Jing Lin\textsuperscript{1}, 
	Zhong Wu\textsuperscript{1}, 
	and Kang Li\textsuperscript{1}
	\thanks{
		Runfeng Zhu, Zhong Wu, Jing Lin, Qingxiang Zhao and Kang Li are with the West China Hospital of Medicine, Sichuan University, Chengdu, China. \\
		Xin Zhong is with the School of Mechanical and Electrical Engineering, University of Electronic Science and Technology of China, Chengdu, China.\\
		Corresponding author: Kang Li. email: {\tt\small likang@wchscu.cn}.
	}
}

\maketitle


{\begin{onecolabstract}
Patients suffering chronic severe pulmonary thromboembolism need Pulmonary Thromboendarterectomy (PTE) to remove the thromb and intima located inside pulmonary artery (PA). During the surgery, a surgeon holds tweezers and a dissector to delicately strip the blockage, but available tools for this surgery are rigid and straight, lacking distal dexterity to access into thin branches of PA. Therefore, this work presents a novel robotized dissector based on concentric push/pull robot (CPPR) structure, enabling entering deep thin branch of tortuous PA. Compared with conventional rigid dissectors, our design characterizes slenderness and dual-segment-bending dexterity. Owing to the hollow and thin-walled structure of the CPPR-based dissector as it has a slender body of 3.5mm in diameter, the central lumen accommodates two channels for irrigation and tip tool, and space for endoscopic camera's signal wire. To provide accurate surgical manipulation, optimization-based kinematics model was established, realizing a 2mm accuracy in positioning the tip tool (60mm length) under open-loop control strategy. As such, with the endoscopic camera, traditional PTE is possible to be upgraded as endoscopic PTE. Basic physic performance of the robotized dissector including stiffness, motion accuracy and maneuverability was evaluated through experiments. Surgery simulation on ex vivo porcine lung also demonstrates its dexterity and notable advantages in PTE.

\def\keywordstitle{Keywords}
\end{onecolabstract}}
 
\begin{multicols}{2}

\section{Introduction}
\subsection{Background}
\begin{figure}
	\centering
	\includegraphics[width=1\linewidth]{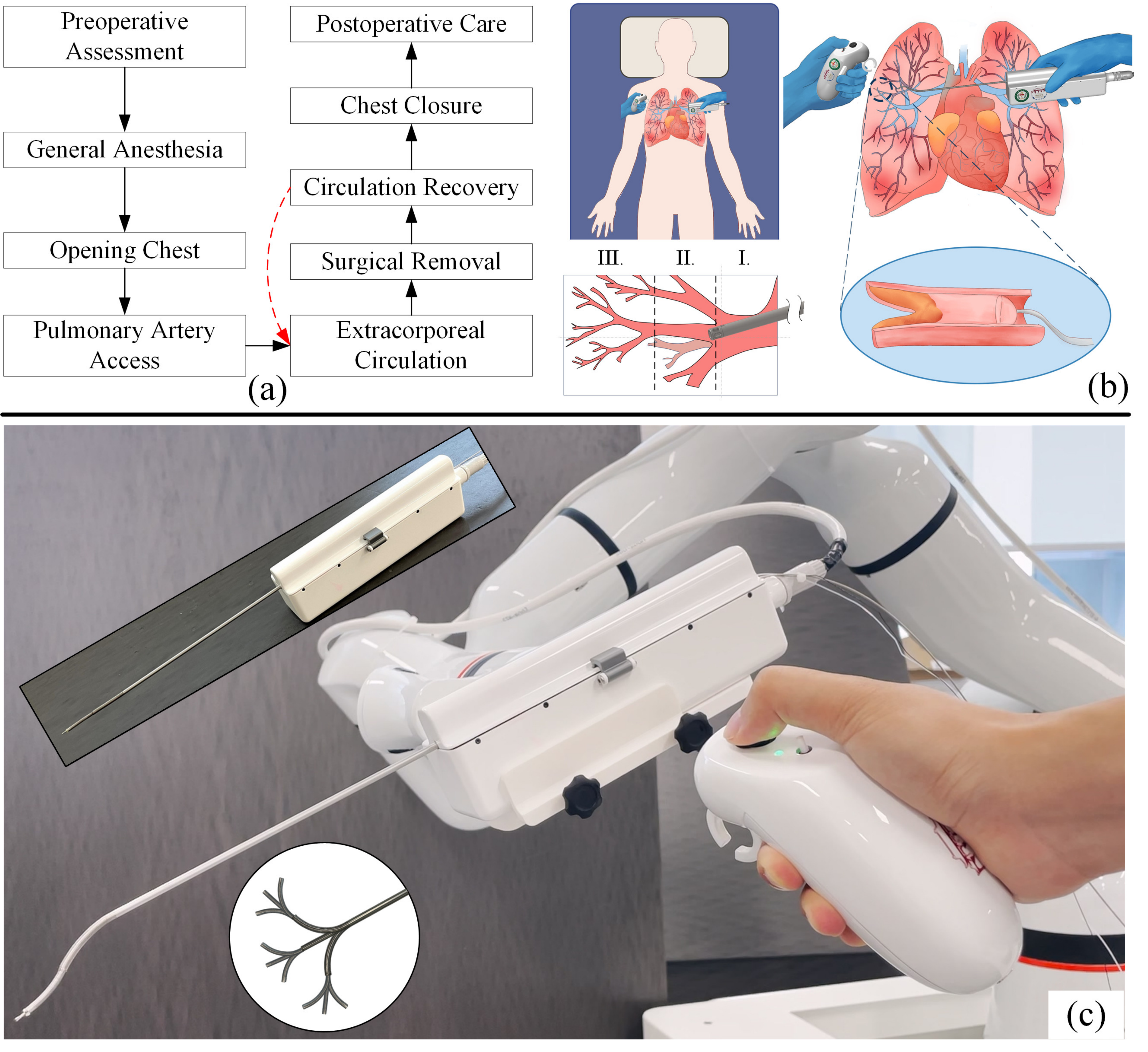}
	\caption{(a) Basic workflow of PTE surgery.  (b) Illustration clots removal procedure. The clots and intima are stripped by an dissector. (c) Robotized steerable dissector enables to enter thin branches of PA easily.}
	\label{fig:pteinstru}
\end{figure}

Because of hypercoagulemia, endocarditis or acquired factors \cite{mcneil2007chronic}, patients suffering chronic thromboembolic pulmonary hypertension (CTEPH) need Pulmonary Thromboendarterectomy (PTE) to remove the thrombus and intima inside pulmonary arteries (PA) \cite{jamieson1998pulmonary, madani2016pulmonary}. The clots in serpentine PA block blood flow, leading to artery hypertension, hemangiectasis, respiration failure and even hemoptysis. Therefore, compared with anticoagulant drugs medication \cite{braams2021evolution}, removing the clots through surgery is more effective in treatment. Figure \ref{fig:pteinstru} (a) shows the basic workflow of this surgery.  All the blood of the patient is aspirated to an extra-corporeal circulation system and patient's circulation arrests for 20 minutes maximally for one turn of surgical removal \cite{jenkins2013surgical}, indicating that patient's circulation shall be restored at the time whether or not the surgical removal has been completed, and another turn of removal starts after the new extra-corporeal circulation is established again \cite{fang2023hyperlactatemia}. Additionally, patient's body temperature is cooled to 20$^\circ$C, which is termed as deep hypothermia cardiac arrest (DHCA) \cite{martin2021pulmonary}. During the surgery, a surgeon holds tweezers and a dissector in the hands \cite{gernhofer2019operative}. The dissector is for stripping the intima, and the tweezers pull clots out. Therefore, effectiveness, safety and completeness are crucial factors in selecting surgical tools \cite{madani2016surgical, guth2022pulmonary}. In the dissector, one aspiration channel is designed to aspire residual blood. However, most existing dissectors for PTE are rigid and straight (Figure 6 in work \cite{madani2016surgical}), which are not able to access into the thin branches of PA easily for lack of dexterity and limited visual field. Additionally, manipulating the straight  dissectors relies on surgeon's experience significantly. Therefore, this work presents a robotized surgical dissector, providing endoscopic visual feedback and dexterity in accessing the deep and thin branches of PA (Figure \ref{fig:pteinstru} (b)). It integrates dexterous robotized slender manipulator, endoscopic camera, illumination, and aspiration channel together to further enhance the surgical outcome (Figure \ref{fig:pteinstru} (c)).
\subsection{Robotic Surgical Devices}
Minimal invasiveness is a hot topic in the development of modern surgeries, and one of the key factors is developing slender and dexterous instruments to reach narrow confined surgical sites through a small port. In recent decades, robotic surgical systems have been well developed to facilitate dexterity and reachability, such as Hugo$^\texttt{TM}$ Robot, Mako Robot \cite{shen2023high} and Mazor X \cite{buza2021robotic}. The robotic systems provide enhanced maneuverability for surgeons. One surgeon could then manipulate multiple robotized instruments to accomplish complicated surgical tasks with endoscopic visual feedback. The systems are generally bulky and need much time for setup in operation rooms. As such, portable  steerable handheld surgical instruments were developed, which have smaller footprint and are cost-friendly \cite{shakir2024clinical}. For example, Hung et. al. \cite{hung2021robotized} developed a handheld smart driller for orthopedic surgery; Yang et. al.  \cite{yang2021design} proposed a handheld surgical device with intuitive manipulability and stable grip force; A handheld printing pen was developed to print stem cells at a chondral wound site \cite{d2016development}. Similar devices were developed for trans-oral surgery \cite{johnson2013demonstration}, neurosurgical micro-dissection \cite{payne2015smart}, endovascular procedures \cite{duan2023technical} and steerable endoscopes \cite{ma2023design}. However, existing handheld surgical tools generally have one or two Degrees of Freedom (DoFs) at the distal section, hindering the application in serpentine PA, because PA presents tortuous shape and clots may locate in the depth of thin PA branches. 

One handheld surgical instrument generally has an ergonomic handheld part, rigid shaft, and steerable section at the tip. The distal steerability is achieved by a continuum manipulator or articulated linkages, while higher degree of steerability (more DoFs) leads to conflict in designing slender surgical manipulator and compact actuation unit. As for continuum robot structures, such as cable-driven or fluid-driven mechanisms, the actuation cables \cite{cao2023design} (or tubes \cite{nguyen2023handheld}) occupy central space. Thus, most slender handheld surgical devices only have one steerable segment for manipulation \cite{cao2023design}. Similarly, cable-driven articulated linkages forming the steerable section also add complication in configuring the cables with the increase of DoFs \cite{lenssen2022sata}. Concentric tube robots (CTR) characterize slender body and thin-walled structure. Nested tubes are compliant with the outer one. As the bending shape relies on interaction among pre-curved tubes \cite{guo2022novel}, this type of handheld surgical tool \cite{xue2022ergonomics} characterizes low stiffness. Kim et al.  \cite{kim2021design} designed a microsurgical robot using delicate Stewart mechanism, with a miniaturized base (diameter of $16$mm), but the dexterity is constrained at the base.  For whichever steerable section, actuators are assembled inside the handheld part, increasing topological complication while more DoFs are needed. Considering the size of PA, the dissector should be miniaturized to below 4mm in diameter and the tip tools like a gripper passes through the central space, which poses significant challenges if using above-mentioned continuum robot structures.
 
Another point in designing a surgical instrument is sterilization convenience to satisfy ISO 10218-1 and ISO/TS 15066 \cite{vossel2021minaro}. For non-robotic surgical devices \cite{riojas2019hand}, this is not a critical issue because pure mechanical hardware is not subject to high temperature (about 130$^\circ$) or oxirane \cite{kim2020development, culmone2021fully}. However, robotic counterparts are sensitive to sterilization due to the attached electronic components. Quick-release mechanism \cite{6853337} is a feasible solution to separate the actuation unit and the instrument unit \cite{lenssen2022sata, de2025cleaning}.

In addition to the structural design, tip tool positioning is also indispensable, which relates to surgery safety.  Accuracy of the tip position directly involves kinematics model.  Zhang et. al. \cite{zhang2025hybrid} proposed a hybrid continuum manipulator and statics model for control. Robust and adaptive control strategy was proposed by Wang et. al. \cite{wang2024robust}. Singh et. al. \cite{singh2026compact} developed a handheld surgical device for minimally invasive neurosurgery, and modeled the tip angle through geometrical mapping. For robotized surgical devices, the model should be computation friendly as the compact actuation unit is not able to provide much space for deploying high-performance electronic hardware.

\subsection{Concentric Push/Pull Robot Arm}
Differed from conventional continuum robot arms, bending of Concentric Push/Pull Robot (CPPR) \cite{oliver2021concentric} is achieved by interaction between two co-axially configured tubes, as shown in Figure \ref{fig:tubebasics} (a). The inner tube and the outer tube are fixed at the tip, and relatively pushing/pulling the inner tube deforms the distal steerable segment \cite{oliver2017concentric} that patterned with slits. The hollow space contributes to passing additional slenderer steerable segments to form multi-segment dexterity. On the basis of previous works \cite{oliver2021concentric, eastwood2016kinetostatic}, we designed tenon-mortise slits along the steerable section, to enhance the overall stiffness in withstanding external loads.

\subsection{Contribution}
As analyzed above, high dexterity and clear visual feedback in the deep branches of thin PA will be beneficial for enhancing surgical outcome and intraoperative effectiveness. Most continuum robotic structures, such as cable-driven/fluid-driven mechanisms and CTRs, may have difficulties in designing a handheld surgical instrument concurrently with an outer diameter to be smaller than 4mm, multi-segment dexterity to adapt with serpentine PA and notable stiffness to manipulate elastic intima, while CPPR has the potential to overcome the limitations. Therefore, in this work, we presented a slender and dexterous dissector to provide an endoscopic PTE strategy, and the key contributions of this work include:
\begin{enumerate}
	\renewcommand{\labelenumi}{\arabic{enumi})}
	\item Designed a compact handheld dissector with a slender body of 3.5mm, 6 DoF of dexterity and notable stiffness to enhance the maneuverability for PTE. With the endoscopic visual feedback, the clots in PA could be dissected more effectively and safely.
	
	\item For the redundant dual-segment CPPR structure, we leveraged constrained optimization to establish the kinematics model, to accurately control the tip position.
\end{enumerate}
The rest of this work is organized as follows. In Section-\ref{Design}, we elaborated on the overall structure of the dissector, after analyzing the specific clinical demands. Kinematics model and analysis about the steerable section are detailed in Section-\ref{Modelling}. Validation about the dissector's basic performance and clinic potential is presented in Section-\ref{Validation}. Finally, Section-\ref{Conclusion} concludes this work.
\section{Robotized Dissector Design}\label{Design}
Based on the 3D morphology of pulmonary arteries\mbox{\cite{huang1996morphometry, singhal1973morphometry}} and existing medication approaches\mbox{\cite{opitz2012technique}}, we analyzed the basic design goals of the dissector following clinical demands{\cite{daily1991dissectors, yoon2019endoscopy}:\\
\texttt{Steerable Section:} The dissector's shape should follow PA's shape to avoid unnecessary trauma while accessing to the thin and deep branches. At least dual-segment bendable shape is required. An endoscopic camera with illumination collects the view inside PA, and irrigation channel is needed to clean the camera. A blood suction channel is required to evacuate the residual blood in PA, which is also the passage to accommodate a gripper.\\
\texttt {Maneuverability:} The dissector must generate sufficient tip force to strip and remove the intima in PA while maintaining accurate tip positioning. Based on clinical demands, the force at the tip should be in the range of 0.5–1.5N to ensure effective tissue manipulation \cite{zhao2010mechanical}.\\	
\texttt {Dimension:} Based on the morphometry of the human pulmonary \cite{huang1996morphometry}, the maximum outer diameter of the tubes should be smaller than 4mm to access into the third-class PA which has a diameter of 4.16 $\pm$ 0.6mm. The total length of steerable section should be 50$\sim$60mm to navigate from the main PA to the third-class PA. For each segment, the bending angle should be larger than 45$^\circ$.\\
\texttt {Weight and Size:} The handheld surgical dissector held by surgeons should be lightweight and small in size\mbox{\cite{lucas2014ergonomics}}, and ergonomics should also be considered.\\
\texttt {Sterilization Convenience:} Quick release and assemble are also needed for sterilization. Integrate no electronic components in the sterilization component.

\begin{table}
	\centering
	\setlength{\tabcolsep}{18pt}
	\caption{Geometric parameters of four tubes forming the two segments (OD: outer diameter, ID: inner diameter).}
	\begin{tabular}{lcccc}
		\toprule
		& \multicolumn{2}{c}{\makebox[1cm][c]{Proximal Segment}} & \multicolumn{2}{c}{\makebox[1cm][c]{Distal Segment}} \\
		\cmidrule{2-5}& \multicolumn{1}{c}{\makebox[1cm][c]{Outer Tube}} & \multicolumn{1}{c}{\makebox[1cm][c]{Inner Tube}} & \multicolumn{1}{c}{\makebox[1cm][c]{Outer Tube}} & \multicolumn{1}{c}{\makebox[1cm][c]{Inner Tube}} \\
		\midrule
		OD (mm) & 3.5   & 3.2   & 2.9   & 2.6 \\
		ID (mm) & 3.3   & 3     & 2.7   & 2.4 \\
		Steerable Section (mm) & 30    & 30    & 20    & 20 \\
		Compliant Section (mm) & 0     & 0     & 40    & 40 \\
		\makebox[2.7cm][l]{Maximum Bending Angle ($^\circ$)} & 50    & 50    & 50    & 50 \\
		\bottomrule
	\end{tabular}%
		\label{tab:Tube_Parameters}%
\end{table}

\subsection{Steerable Section Selection}
\begin{figure}
	\centering
	\includegraphics[width=1\linewidth]{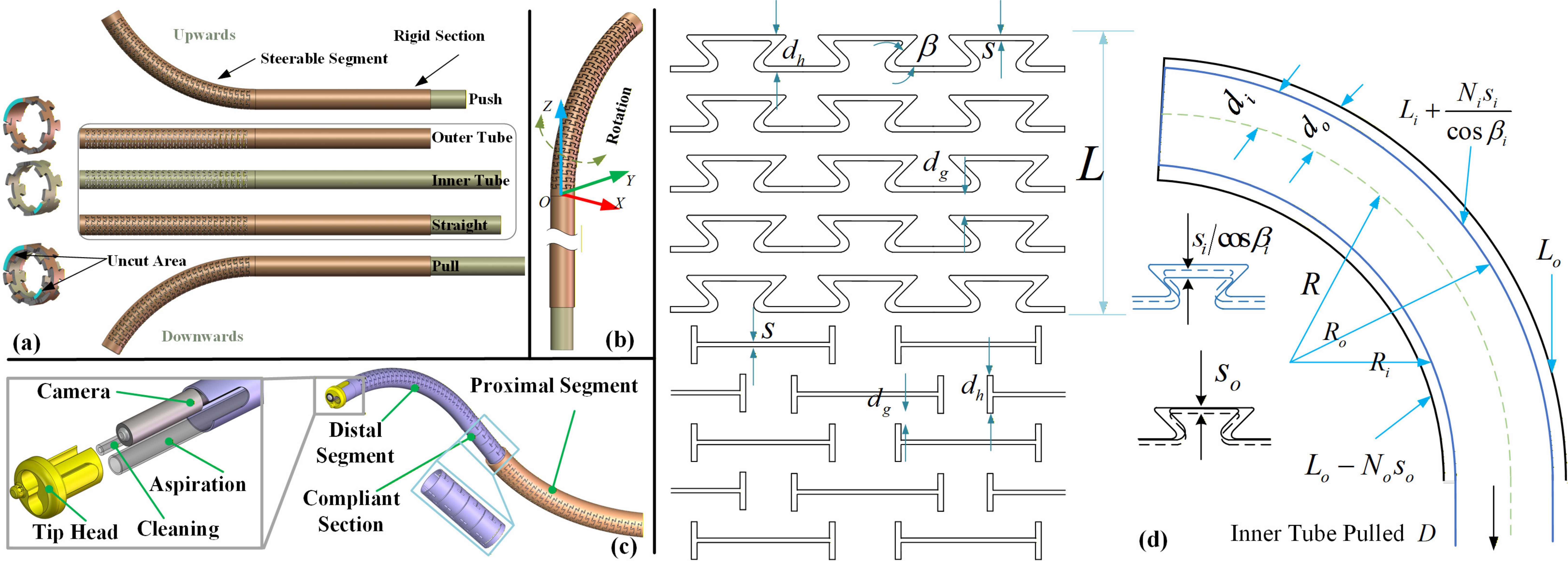}
	\caption{Basic working principle of one CPPR manipulator. (a) One steerable segment consists of two patterned tubes and are assembled with the patterned section oppositely configured. (b) One rotation DoF enables 3D bending. (c) Distal segment passes through the central hollow lumen of the proximal segment. Tip head integrates an endoscopic camera, irrigation passage and an instrument channel which can also be used for suctioning residual blood. (d) Tenon-mortise slits on the inner tube and outer tubes are well designed. ‘I’-shaped slits are designed on the distal segment, which overlap with the steerable section of the proximal segment. The shape of the 'I'-shaped slits section follows the shape of proximal segment.}
	\label{fig:tubebasics}
\end{figure}
As shown in Figure \ref{fig:tubebasics} (a), one steerable segment was developed based on concentric push/pull tubular robot arm \cite{9536967}, which consists of a pair of hollow tubes. Each tube was manufactured with dense tenon-mortise slits using laser cutting technology. Then the two tubes are fixed at the tip through laser welding, and the patterned sections are oppositely configured. Consequently, relatively pushing/pulling the inner tube (the outer tube is assumed fixed) at the rear end generates planar bending. The main reason why we specifically select tenon-mortise slits is that tenons and mortises could gradually interlock together to enhance overall stiffness with the increase of bending angle, which presents notable advantages compared with square slits \cite{lee2015anisotropic} and diamond-shaped patterns \cite{luo2021design}.  

Since higher dexterity is desired, we attached a rotation DoF at the rigid section, providing 3D task space for the whole steerable segment, as shown in Figure \ref{fig:tubebasics} (b). Another advantage is the thin-walled structure (the wall thickness of each tube can be miniaturized to 0.1mm), so much central hollow space could be employed to pass a slenderer steerable segment or a tip tool, as shown in Figure \ref{fig:tubebasics} (c). After analyzing the clinical demands, we finalized that a dual-segment steerable section is needed to enter the depth of PA and their basic geometric parameters are summarized in Table \ref{tab:Tube_Parameters}. To enhance the overall stiffness, we require that the tenons and mortises on the two tube should interlock together simultaneously, as shown in Figure. \ref{fig:tubebasics} (d). The basic parameters of the slits consist of: slit width $s$, slit gap distance $d_g$, slit height $d_h$, tilted angle $\beta$ and number of the slits $N$. $L$ denotes the length of the patterned section. While the segment bends maximally, the tenon-mortise slits on the inner tube extend and those on the outer tube compress. Since the two tubes share an identical bending curvature and the shape can be assumed as an arch, we can require:
\begin{equation}
	\begin{gathered}
		{L_o} - {N_o}{s_o} = (R - {d_o})\theta  \hfill \\
		{L_i} + \frac{{{N_i}{s_i}}}{{\cos {\beta _i}}} = (R + {d_i})\theta  \hfill \\
		{L_o} = (R + {d_o})\theta  \hfill \\ 
	\end{gathered} 
\end{equation}
where subscripts $o$ and $i$ respectively denote the parameters for the outer tube and the inner tube. $R$ is bending radius and $\theta$ is the bending angle (central angle of the arch). $d$ is the distance measuring from the backbone to the tube wall's center line. Therefore, we can obtain the relationship between the two tubes' slits:
\begin{equation}
 {L_o} - {L_i} - \frac{{{N_i}{s_i}}}{{\cos {\beta _i}}} = ({d_o} - {d_i})\frac{{{N_o}{s_o}}}{{2{d_o}}}
\end{equation}
The length of the steerable section and basic diameters of the tubes are predefined following design goals. In addition, slit height and gap distance are also related to the length $L$:
\begin{equation}
	N(d_h+d_g)=L
\end{equation}
Therefore, the slits' parameters can be defined and accordingly computed, which are listed in Table \ref{tab:GeometricSlits}.
 \begin{table}[]
	\centering
	\caption{Geometric parameters of the slits for the tubes.}
	\label{tab:GeometricSlits}
	\begin{tabular}{lllll}
		\hline
		& \multicolumn{2}{c}{Proximal Segment} & \multicolumn{2}{c}{Distal Segment} \\ \cline{2-5} 
		& Outer Tube        & Inner Tube       & Outer Tube       & Inner Tube      \\ \hline
		$\beta$ ($^\circ$) & 63                & 69               & 63.6             & 70              \\
		$s$(mm)    & 0.03              & 0.05             & 0.03             & 0.049           \\
		$d_g$(mm)   & 0.56              & 0.39             & 0.31             & 0.32            \\
		$d_h$(mm)   & 0.25              & 0.25             & 0.25             & 0.25            \\
		$N$    & 50                & 68               & 57               & 83              \\ \hline
	\end{tabular}
\end{table}
Since the compliant section of distal segment should passively follow the shape of the proximal segment, we designed 'I'-shaped slits for passive omni-directional bending. The length and width of the 'I'-shaped slits are identical with the tube's tenon-mortise slits, which ensures that the stiffness of the passive compliant section is smaller than the stiffness of the proximal segment. The bending angles in design goal are not too high (only more than $50^\circ$), requiring small elastic range, so we chose medical steel (316L) as the tube material. Although Nitinol is a better alternative for its large elastic range, it is relatively high-cost and characterizes low stiffness (30GPa of Young's Elasticity Modulus).

As shown in Figure \ref{fig:stiffnessanalysis} (a), one CPPR segment can be assumed as a cantilever beam, the ability to withstand external load relates to the stiffness. For the beam, the stiffness can be mathematically expressed by $EI$, where $E$ is the Young's Elasticity Modulus and $I$ is the second moment of inertia. Larger $I$ denotes higher stiffness. Figure \ref{fig:stiffnessanalysis} (b) shows the cross section of one tenon-mortise slit, and the second moment of inertial can be related to the central angle $\alpha$ of the uncut area and tube's size:
\begin{equation}
	I = \frac{1}{8}(\alpha  - \sin \alpha )(R_{OD}^4 - R_{ID}^4)
\end{equation}
where $R_{OD}$ and $R_{ID}$ respectively denote the outer diameter and inner diameter (Figure \ref{fig:stiffnessanalysis} (b)). While it bends maximally, tenon and mortise interlock together. As shown in Figure \ref{fig:stiffnessanalysis} (c) (highlighted in red circles), the interlock means the increase of $I$. On the contrary, second moment of inertia of CPPR segment with square slits in bending status is lower (Figure \ref{fig:stiffnessanalysis} (c)). With a load of $F$, the deflection $w$ can be then solved by:
\begin{equation}
	w = \frac{{F{L^3}}}{{3EI}}
\end{equation}
where $L$ is the length. Because the tenon-mortise slits characterize larger $I$, the deflection of it will be lower than that of square slits design.
\begin{figure}
	\centering
	\includegraphics[width=1\linewidth]{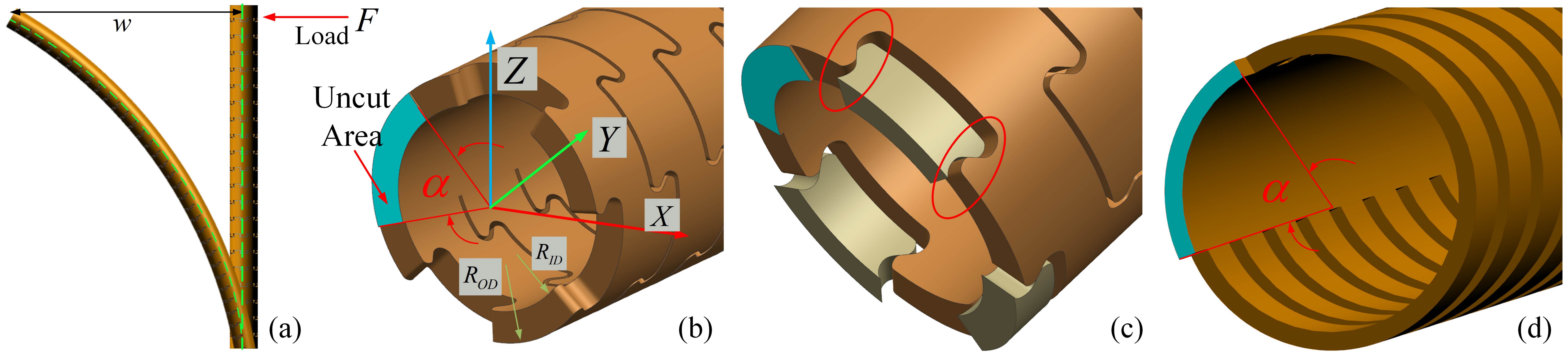}
	\caption{(a) One CPPR steerable segment is assumed as a cantilever beam, and deflection $w$ reflects the capability of withstanding external load $F$. (b) Cross section of one tenon-mortise slit, and the stiffness is mainly defined by the central angle $\alpha$ of the uncut area and tube's size. (c) Cross section of one tenon-mortise slit in maximum bending or in with-load status, where tenon and mortise interlock together to increase the second moment of inertia. (d) Cross section of square slit. It also has an uncut area parameterized by $\alpha$, but the stiffness is relatively lower compared with tenon-mortise slit design. }
	\label{fig:stiffnessanalysis}
\end{figure}

At the tip of the distal segment, a 3D-printed tip disk made by nano printing technology (nanoArch S140, BMF, China) is assembled, integrating an endoscopic camera (OCHTA10-KL1C, Omnivision, USA) with a diameter of 1.3mm. Two channels with diameters of 0.8mm and 1.2mm  were respectively designed for irrigation and aspiring the residual blood in PA, as illustrated in Figure \ref{fig:tubebasics} (c). The aspiration could also be utilized to accommodate a tool, like a gripper. The tip head is smooth to avoid scratching fragile PA.

\subsection{Mechanical and Electronic Design}
\begin{figure*}
	\centering
	\includegraphics[width=1\linewidth]{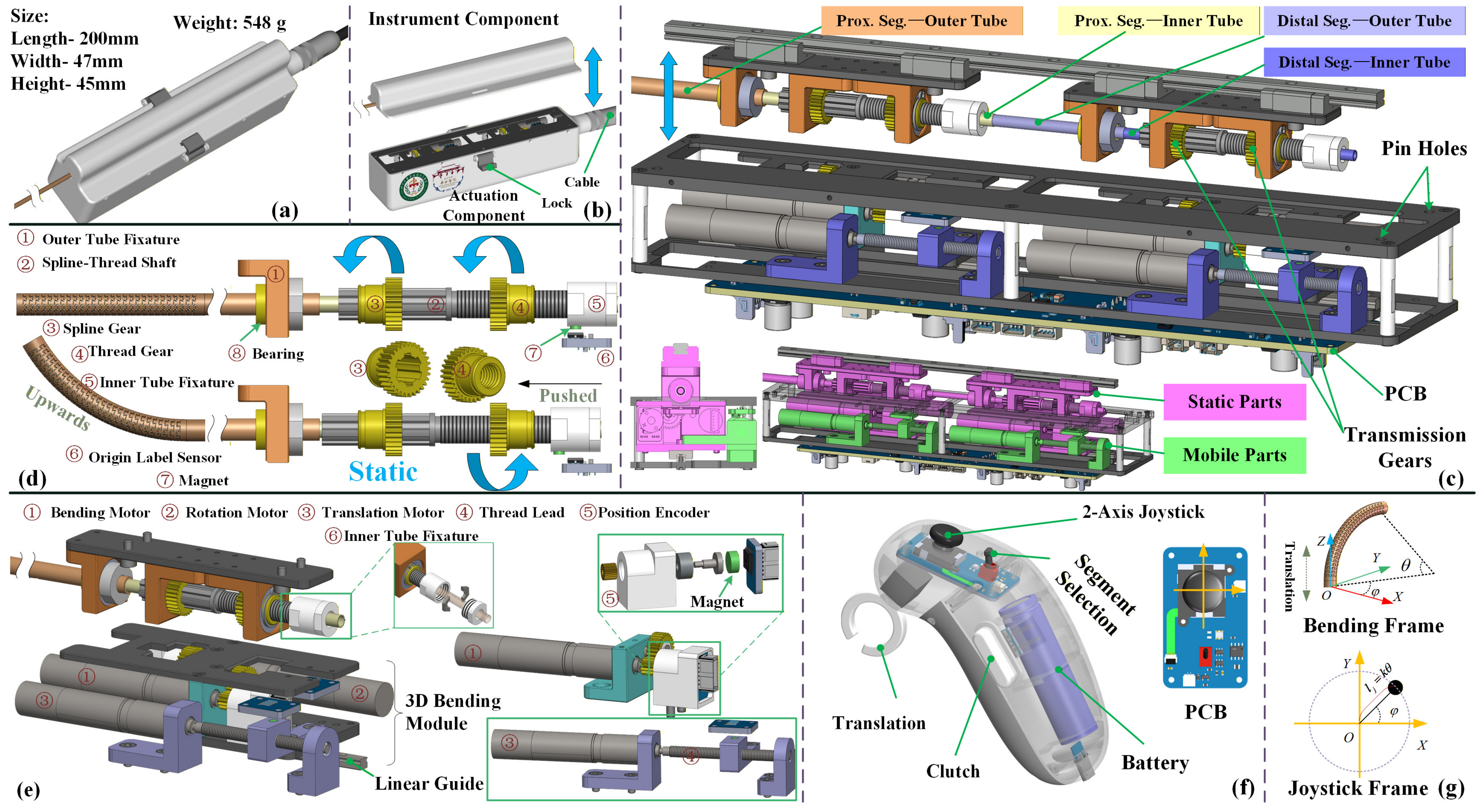}
	\caption{Overall Design of the dissector. (a) Handheld part. (b) Quick release and assemble for sterilization convenience. (c) Instrument component and actuation component. Static and mobile parts are highlighted. (d) Basic working principle of actuating a single segment with compact module. (e) Actuator design for a single segment. (f) Wireless control unit for teleoperation. (g) Coordination between joystick frame and bending frame.}
	\label{fig:design}
\end{figure*}
In addition to the flexible manipulator, designing proper actuation unit is also indispensable. Figure \ref{fig:design} (a) shows the handheld part of the dissector, consisting of an instrument component and an actuation component. No electronic hardware was integrated in the instrument component for sterilization convenience. As shown in Figure \ref{fig:design} (b), two locks bond the two components for quick release and assembly. A cable is set on the actuation component to provide command signal and power source. Each segment has three DoFs, i.e., axial translation, bending and rotation. All actuation modules are arranged inside the actuation component, and the tubes (Figure \ref{fig:tubebasics} (c)) are integrated in the instrument component. As shown in Figure \ref{fig:design} (c), three DC motors (DCU08017P12, Moons, China) whose power, velocity and torque were carefully calculated, are compactly arranged in the actuation component, and gears ensure precise rotation transmission between the two detachable components. 

 For bending and rotation DoFs, we have specifically designed an actuation mechanism, as shown in Figure \ref{fig:design} (d). The bending is achieved by pushing/pulling the inner tube within a short distance (maximally 5mm). A spline-thread shaft is proposed for compact actuation, which is motivated from keyed lead screw structure \cite{rox2020mechatronic} and roller gear mechanism \cite{girerd2020design, morimoto2017design}. The outer tube is fixed on a bearing, so it only has a rotation DoF. The spline-thread shaft is fixed with the inner tube, such that pushing/pulling the shaft generates bending. In addition, the two tubes are also constrained to relative translation, and rotating the inner tube will rotate the whole segment, as illustrated in Figure \ref{fig:design} (e). Two gears were respectively assembled on the thread section and spline section, as shown in Figure \ref{fig:design} (d). Once the spline gear keeps static, thread gear rotates to translate the spline-thread shaft and further generates push/pull distance to the inner tube. Similarly, concurrent rotation (equal velocity and angular displacement) of the two gears only generates rotation along the shaft and generates rotation DoF for the steerable segment. Figure \ref{fig:design} (e) shows the detailed design of one actuation module. Beside the built-in encoder in the motors, we have also designed hall sensors (DRV5055A4QDBZR, TI) to label the origin of the translation DoF and magnetic encoders (AS5048B, AMS) to track the real-time angular displacement of the motors. The axial translation DoF is achieved by a lead screw guide, so rotation of motor \Circled{3} translates the whole module axially. The overall dimension of the handheld dissector is $200$mm$\times 47$mm$\times 45$mm, and the total weight of the dissector is 548g.

To enable tele-operation, we separate the control unit from dissector, which is different from common handheld surgical tools' design \cite{wang2022hybrid, chitalia2020design}. This is also beneficial for actuator miniaturization. The wireless control unit sends command to actuators via WiFi telecommunication. As shown in Figure \ref{fig:design} (g), a two-axis joystick controls the bending angle and direction angle of a single segment, and segment selection switch determines which segment to steer. For the first mode, joystick frame and bending frame are mapped. Namely, the joystick angle is equal to the direction angle $\varphi$, and the distance $l_j$ is proportional to bending angle $\theta$. This is able to directly control the shape of the each segment, so we also hope to position the tip using the wireless joystick, i.e., the tip position $(x,y,z)$ could be directly controlled by the surgeon, which requires kinematics model.

\section{Modeling}\label{Modelling}
In this section, we modeled the mapping between push/pull distance and the robot shape configuration, based on which the tip pose and actuation inputs are mapped to build kinematics.

\begin{figure}
	\centering
	\includegraphics[width=1\linewidth]{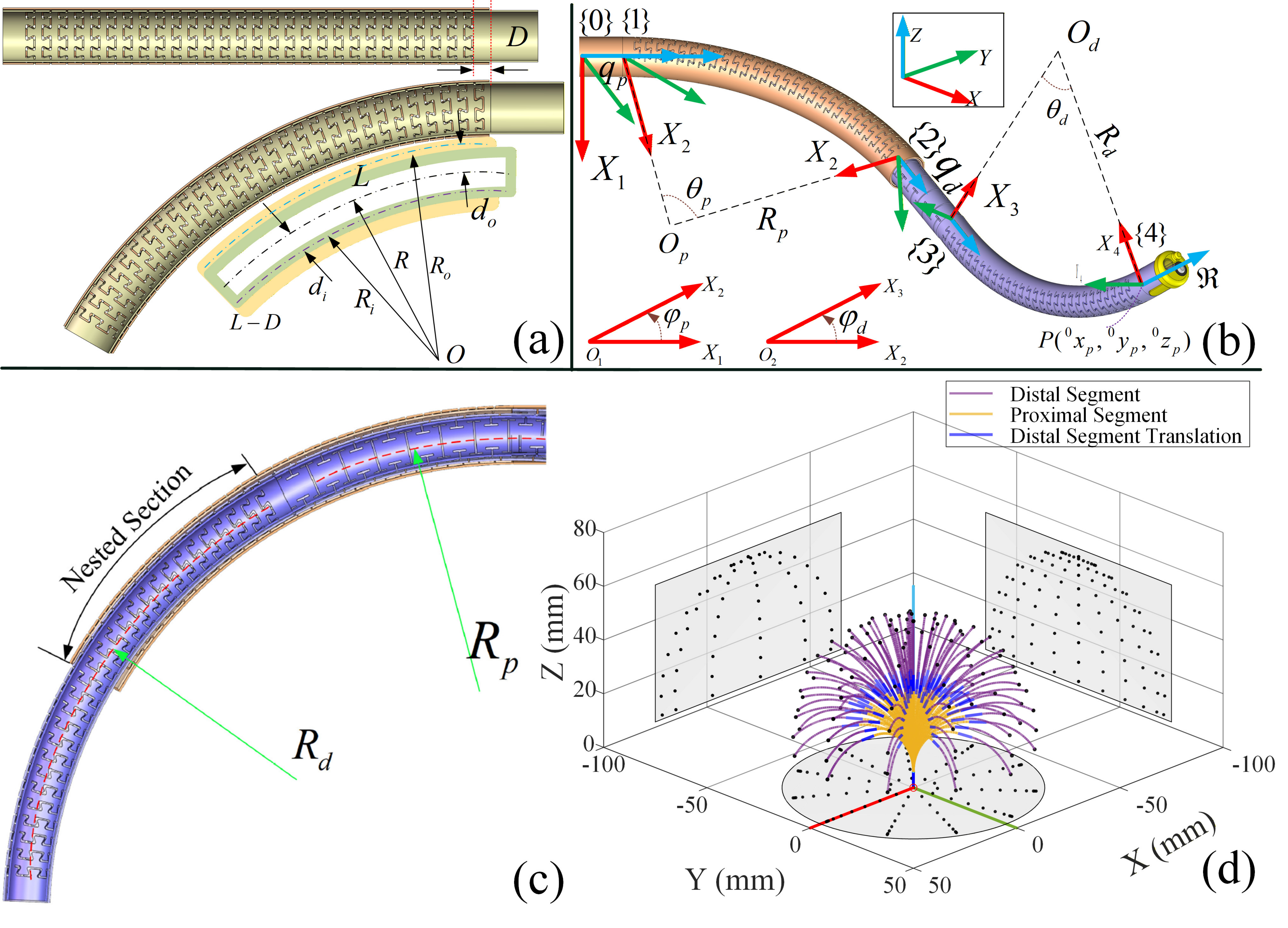}
	\caption{(a) Bending angle $\theta $ with pulling distance $D$ of a single manipulator. (b) The dual-segment CPPR manipulator's shape can be accordingly solved, and the tip pose w.r.t the base frame is computed via the shape. (c) The distal segment is nested nest into the proximal segment, but their bending curvature should be identical to avoid mutual interaction and even breakage. (d) Task space illustration.}
	\label{fig:model}
\end{figure}
\subsection{Bending of a Single Segment}
Figure \ref{fig:model} (a) shows the straight and deformed shapes of a single segment. It should be noted that only pulling action is considered for brevity because the rotation DoF could bring the steerable segment to any bending plane. Then, the deformed shape is solved under following assumptions: 1) The backbone is a constant-curvature arch as the slits are densely distributed (0.25mm gap distance), where both inner tube and the outer tube share an identical arch center $O$; and 2) The space (0.05mm) between outer surface of inner tube and inner surface of outer tube is neglected.

As shown in Figure \ref{fig:model} (a), the outer tube is fixed and pulling distance acting on the inner tube deforms the manipulator. The length of the steerable section is denoted as $L$, and accordingly the arch length of inner tube is $L-D$. The steerable section's length of the outer tube maintains a constant value $L$. Since they have an equal bending angle $\theta$, we can obtain: 
\begin{equation}\label{eq:D2theta}
	\theta  = \frac{L}{{R + {d_o}}} = \frac{{L - D}}{{R - {d_i}}},
\end{equation}
where $d_o$ and $d_i$ respectively denote the distance from backbone to tube wall's center line of outer tube and inner tube, and $R$ is bending radius shown in Figure \ref{fig:model} (a). In this work, we assume the inextensible layer lies at the center of the tube wall. Therefore, the mapping between pulling distance $D$  acting on the inner tube and the bending angle $\theta$ is:
\begin{equation}
	\theta  = \frac{D}{{{d_o} + {d_i}}}.
\end{equation}
The linear mapping is beneficial for solving the inverse process.

\subsection{Kinematics}
\subsubsection{Forward Kinematics}
Figure \ref{fig:model} (b) shows the coordinate frames and deformed shape of the robot arm. Forward kinematics model maps between tip pose and actuation inputs $Q= \{ {q_p},{D_p},{\varphi _p},q_d,{D_d},{\varphi _d}\} $, where $q_p$ and $q_d$ respectively denote the translation displacements of the proximal segment and the distal. In detail, the tip pose (frame $\{4\}$) w.r.t. the global base frame $\{0\}$ should be calculated with the actuation configuration $Q$. $D_p$ and $D_d$ are the pulling distance acting on the proximal segment and the distal, respectively. Accordingly, ${\varphi _p}$ and ${\varphi _d}$ are direction angles of the two segments. Frame $\{0\}$ is deemed as robot base. It should be noted that the steerable section of the distal segment could also be nested into the proximal segment (Figure \ref{fig:model} (c)). In this scenario, the curvature of the two segments should be equal, otherwise the mutual interaction will cause breakage at the distal segment. The requirement is mathematically set as: $\varphi_d=0$ or $\varphi_d=\pi$ and ($R_p=R_d$), if ${q_d} < 0$. Herein, the bending shape of the segments is assumed as an arch, so the forward kinematics could be established via piecewise constant curvature assumption \cite{webster2010design, zhao2024controller}, with which the tip pose could be solved via Homogeneous Transformation. This is sophisticated and abbreviated as: 
\begin{equation} \label{FK}
	{P}=f(Q), {\Re}=g(Q)
\end{equation}
where $P(^0x_p,^0y_p,^0z_p)$ is the tip position and $\Re$ is the orientation. With \eqref{eq:D2theta}, the bending angle of each segment could be mapped with the pulling distance. Because the tip position and orientation are highly coupled, only position vector ${}_0^4{P} \in {R^3}$ and one orientation ${}_0^4{\Re} {\rm{ = }}\overrightarrow {{O_4}{Z_{\rm{4}}}} $ are considered in modeling, so the to-be-controlled tip pose is ${\Theta} {\rm{ = }}\left[ {\begin{array}{*{20}{c}}
		{{}_0^4{P}}&{{}_0^4{\Re} }
\end{array}} \right]$.
 For ease of computation, orientation ${\Re} \in {R^3}$ is denoted by an unit vector:
 unit vector:	${\Re}  = {}_0^4{T}{\left[ {\begin{array}{*{20}{c}}	0&0&1&1	\end{array}} \right]^T}-(^0x_p,^0y_p,^0z_p)$. Figure \ref{fig:model} (c) shows the task space of the end effector, which covers a volume of around 95.1$\text{cm}^3$.
\subsubsection{Inverse Kinematics}

Accordingly, once a desired task configuration desired ${\tilde \Theta}{\rm{=}}\left[{\begin{array}{*{20}{c}}{{\tilde P}}&{{\tilde \Re }}\end{array}} \right]$ is given, an actuation inputs ${\hat Q }$ should be quickly and accurately solved. ${\tilde P}$ denotes the target tip position and ${\tilde \Re}$ is the target orientation vector. Because of non-linearity and hyper redundancy, the optimal actuation inputs ${\hat Q}$ are solved via constrained optimization algorithm, and the object function is:
\begin{equation}\label{Loss_Fun}
	{\hat Q} = \arg \min \left\| {f({\hat Q}) - {\tilde P}} \right\| + \chi \left\| {g({\hat Q}) - {\tilde \Re }} \right\|
\end{equation}
where  $f({\hat Q})$ and $g({\hat Q})$ are both from forward kinematics to map the actuation inputs with tip position and orientation, respectively. A weighting coefficient $\chi  \in \left( {0,1} \right)$ tunes the orientation item. While the robot arm could not reach the orientation, $\chi$ is then set to zero and only the position ${\tilde P}$ is the goal to reach. The actuation inputs should also be limited within given range:
\begin{equation}\label{Opt_Con}
	\left\{ {\begin{array}{*{20}{l}}
			{0 \le {q_p} \le {q_{p,\max }},{q_{d,\min }} \le {q_d} \le {q_{d,\max }}}\\
			{0 \le {\varphi _p} \le 2\pi ,0 \le {D_p},{D_d} \le {D_{\max }}}\\
			{{\varphi _d} = 0, R_p=R_d,  {\rm{if}}  {q_d} < 0}\\
			{0 \le {\varphi _d} \le 2\pi, {\rm{if}}  { q_d} \ge 0}
	\end{array}} \right.
\end{equation}
where the first inequation limits the translation DoFs, and the second limits the rotation angle of the proximal segment and bending angles. The range of $\varphi_d$ is conditional with $q_d$, which is difficult to solve the inverse, so we revise it as:
\begin{equation}
	0 \le {\varphi _d} \le [{\rm{sign}}({q_d}) + 1]\pi 
\end{equation}
where ${\rm{sign}}(\cdot)$ is sign function. To solve the optimization problem, we leveraged Newton-Raphson algorithm, and the constraints are added into object function:
\begin{equation}
	\begin{split}
		H({Q}) &= \left\| {f({\hat Q}) - {\tilde P}} \right\| + \chi \left\| {g({\hat Q}) - {\tilde \Re} } \right\| \\
		&\quad - {\mu _i}\sum\limits_{i = 1}^M {{h_i}({Q})} 
	\end{split}
\end{equation}
where $\mu_i=0.1$ is the penalty coefficient for the $i{\rm{th}}$ constraint function, and $M$ is the total number of constraints (equation \eqref{Opt_Con}). Iterative approach was used to solve the optimization problem, and the actuation input $\hat Q$ in the $n{\rm{th}}$ iteration step is:
\begin{equation} \label{Newton-Rapshon}
	{Q_n} = {Q_{n - 1}} - \frac{{H({Q_{n-1}})}}{{J({{Q_{n - 1}}})}}
\end{equation}
where $J({Q_n})$ is the Jacobian of $H(\cdot)$ in the $n{\rm{th}}$ iteration.
\subsection{Singularity Analysis}
In robotics, Jacobian matrix can be utilized to analyze singularity. Between adjacent control instances, both end effector and actuation inputs vary slightly and Jacobian matrix can be obtained as:
\begin{equation}
	\left[ {\begin{array}{*{20}{c}}
			{{\delta P}}\\
			{{\delta \Re }}
	\end{array}} \right] = \left[ {\begin{array}{*{20}{c}}
			{{J_P}}\\
			{{J_\Re }}
	\end{array}} \right]{\delta Q}.
\end{equation}
While robot moves within small range between adjacent instances $k{\rm{th}}$ and $(k+1){\rm{th}}$, this could also be linearized as:
\begin{equation}\label{Linearize_Jac}
	\begin{array}{l}
		{P}(k + 1) - {P}(k) = {{J_P}}(k)\left[ {{Q}(k + 1) - {Q}(k)} \right]\\
		{\Re} (k + 1) - {\Re} (k) = {J_\Re }(k)\left[ {{Q}(k + 1) - {Q}(k)} \right]
	\end{array}
\end{equation}

Although we could use the proposed inverse kinematics to find the optimal actuation inputs for a given desired task configuration $ {\tilde \Theta}$, it could not help to detect singularity. In \eqref{Linearize_Jac}, the actuation inputs could be revised as: 
\begin{equation}
	\begin{split}
		{Q}(k + 1) &= {Q}(k) + \left[ {{J}_{_\Theta }^T(k){{J}_\Theta }(k)} \right]^{{\rm{ - 1}}} {J}_{_\Theta }^T(k) \\
		&\quad \times \left[ {{\tilde \Theta} (k + 1) - {\Theta} (k)} \right]
	\end{split}
\end{equation}
Once $ \left[ {{J}_{_\Theta }^T(k){{J}_\Theta }(k)} \right]$ is not of full rank, it can not be inverted, which means that singularity presents.
\section{Experimental Validation}\label{Validation}
In this section, basic performance of the dissector was first validated, including dexterous motion, kinematics model and stiffness comparison. Then, ex vivo tests were conducted to validate its clinic feasibility. 
\begin{figure*}
	\centering
	\includegraphics[width=1\linewidth]{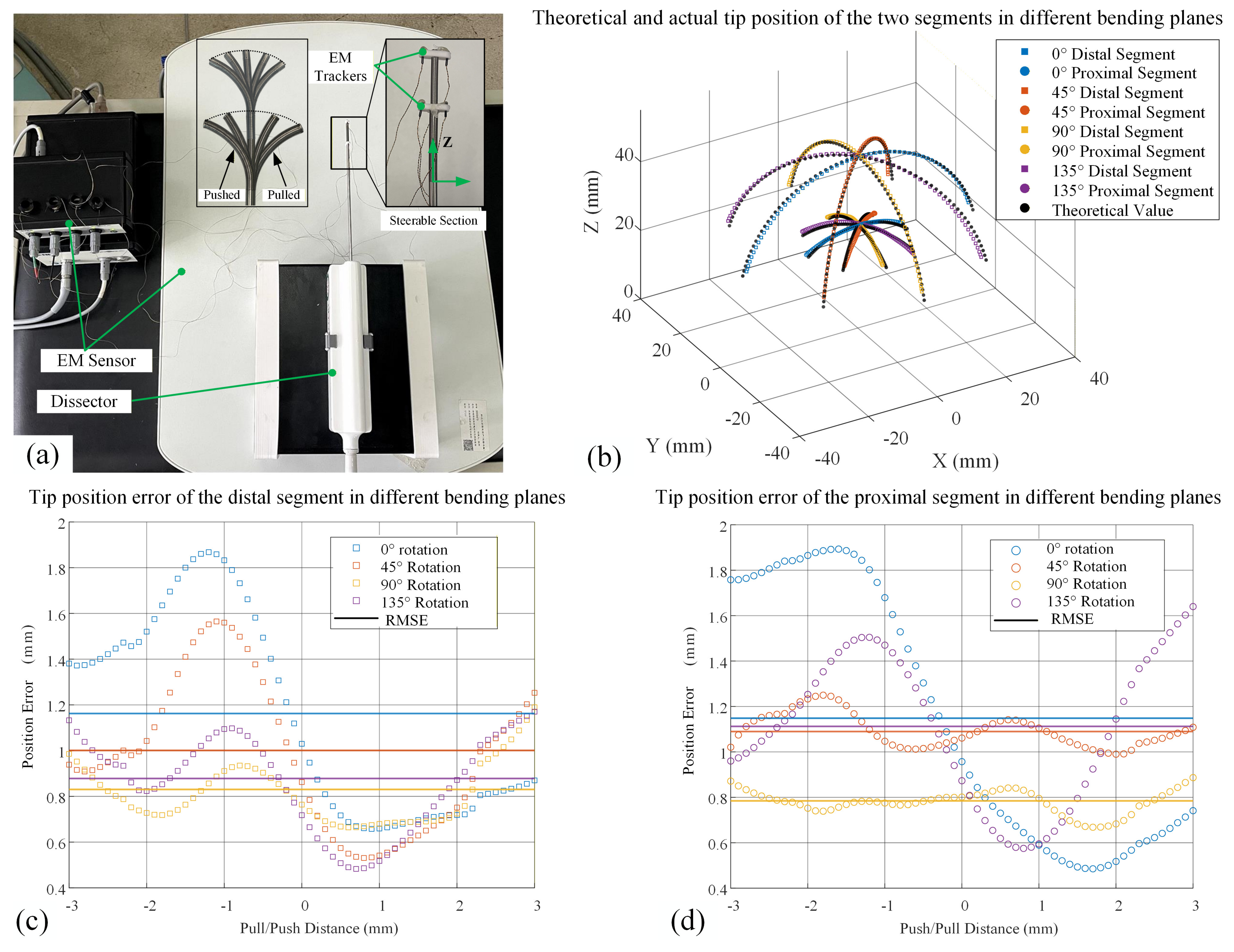}
	\caption{(a) Experiment setup to validate the forward kinematics model. (b) Comparison between actual tip position and the theoretical values for the two segments in eight different bending planes. (c) Position error of the distal segment in different actuation configurations and four bending planes. (d) Position error of the proximal segment in different actuation configurations and four bending planes.}
	\label{fig:fkval}
\end{figure*}

\subsection{Robot Kinematics Validation}
Accuracy of kinematics model is crucial for tip pose control and master-salve manipulation. The forward kinematics model was first validated by comparing the actual tip position and the theoretical values.
\subsubsection{Forward Kinematics Validation}
Figure \ref{fig:fkval} (a) shows the experiment setup to measure the actual tip position of each segment in different actuation inputs. The dissector was fixed on a sensing system (NDI, AURORA, Canada). Two magnetic trackers were glued beside the base and tip of the steerable section to collect the actual position. Since the trackers are not able to be installed at the base point and the tip point (inside the tubes), using the mean value of the two trackers is a feasible solution. Because the length of each tracker is around 5mm, and we just need a point position. We calibrated the distance between the two groups of trackers to minimize the measurement error. In detail, robot manipulator was actuated with multiple shapes, and the deviations between the sensed tip position and the actual position (checkerboard made of coordinate worksheet) was used to adjust the installation position of the trackers.
The inner tube was actuated from -3mm to 3mm (minus: pulled, positive: pushed) with an increment of 0.1mm. In addition, the rotation DoF was also activated to collect the tip points in whole task space, i.e., the tip points in bending planes of $0^\circ, 45^\circ, 90^\circ$ and $135^\circ$ were sampled and compared. As can be observed from Figure \ref{fig:fkval} (b), the actual tip position is close to the theoretical values. Figure \ref{fig:fkval} (c) and (d) respectively show the position error of the distal segment and the proximal segment, including the point-by-point position error and the root mean square error (RMSE). The maximum position error is around 1.85mm and the maximum RMSE in the eight settings is around 1.2mm. The error is from two aspects: 1) The rotation DoF essentially has discrepancy; 2) The robot shape in large bending status does not necessarily follow a perfect arch.

\subsubsection{Inverse Kinematics Validation}
The proposed optimization algorithm to find the optimal actuation inputs for a given desired tip pose is investigated in terms of: accuracy and computation time consumption. Initially, the value of $\chi$ in \eqref{Loss_Fun} was set to 0.5, and the maximum iteration step in \eqref{Newton-Rapshon} was set to 20. While the objective function was not able to converge to zero, two scenarios occur: 1) the tip could reach the given position but could not satisfy the orientation requirement, and 2) the given desired tip position is not within task space. In this experiment, we did not consider the second scenario as the given desired tip configuration is always properly given. Therefore, once the objective function \eqref{Loss_Fun} could not converge, $\chi$ was then set to zero to only meet position requirement. 

We set a spiral path for robot tip following, and there were totally 12 task points to reach, and the orientation followed a random unit vector.  This simulation only validates the accuracy and time cost of the optimization-based inverse kinematics model. Figure \ref{fig:ikvalidation} (a) shows the task points, robot tip position and the shapes. The constraints in \eqref{Opt_Con} were set as: ${{q_{p,\max }}}={{q_{d,\max }}}=$ $10$mm, ${{q_{d,\min }}}=$ $-5$mm and ${{D_{\max }}}=$ $5$mm. All the end effectors reached the desired points, and the shapes present multiple configurations to map with the task points. The calculated RMSE between task points and robot tip points was 0.09mm, and the maximum distance was 0.23mm. If the robot could simultaneously reach the tip and orientation (value of the object function is below than 0.01), which means that system just needs to search the optimal actuation inputs ($\chi=0.5$), the time consumption was around 0.4s (MATLAB 2021a) in each control instance. While it failed to satisfy the orientation requirement, system needs to calculate again, i.e. change $\chi=0.5$ into $\chi=0$, and the time consumption was around 0.8s. Figure \ref{fig:ikvalidation} (b) shows the actuation inputs of each DoF, all of which are all within given actuation range.

Then, the robot arm was commanded to follow a circular path consisting of 200 tip points. Figure \ref{fig:ikvalidation} (c) shows the comparison between the actual path and the desired path, and the position error in each step is shown in Figure \ref{fig:ikvalidation} (d). The maximum error is around 2mm, and RMSE is around 1.1mm. This is consistent with the error of forward kinematics. As recorded in the Supplementary Video, the whole motion process is smooth. The path was then followed three times, and the results are shown in Figure \ref{fig:ikvalidation} (e) and (f). The maximum error was also around 2mm. Compared with work \cite{zhang2022survey} which is a survey of continuum robotics, our model has an accuracy of 2mm with 60mm of length. The results prove that our robot arm has a reliable and good path-following performance.
\begin{figure}
	\centering
	\includegraphics[width=1\linewidth]{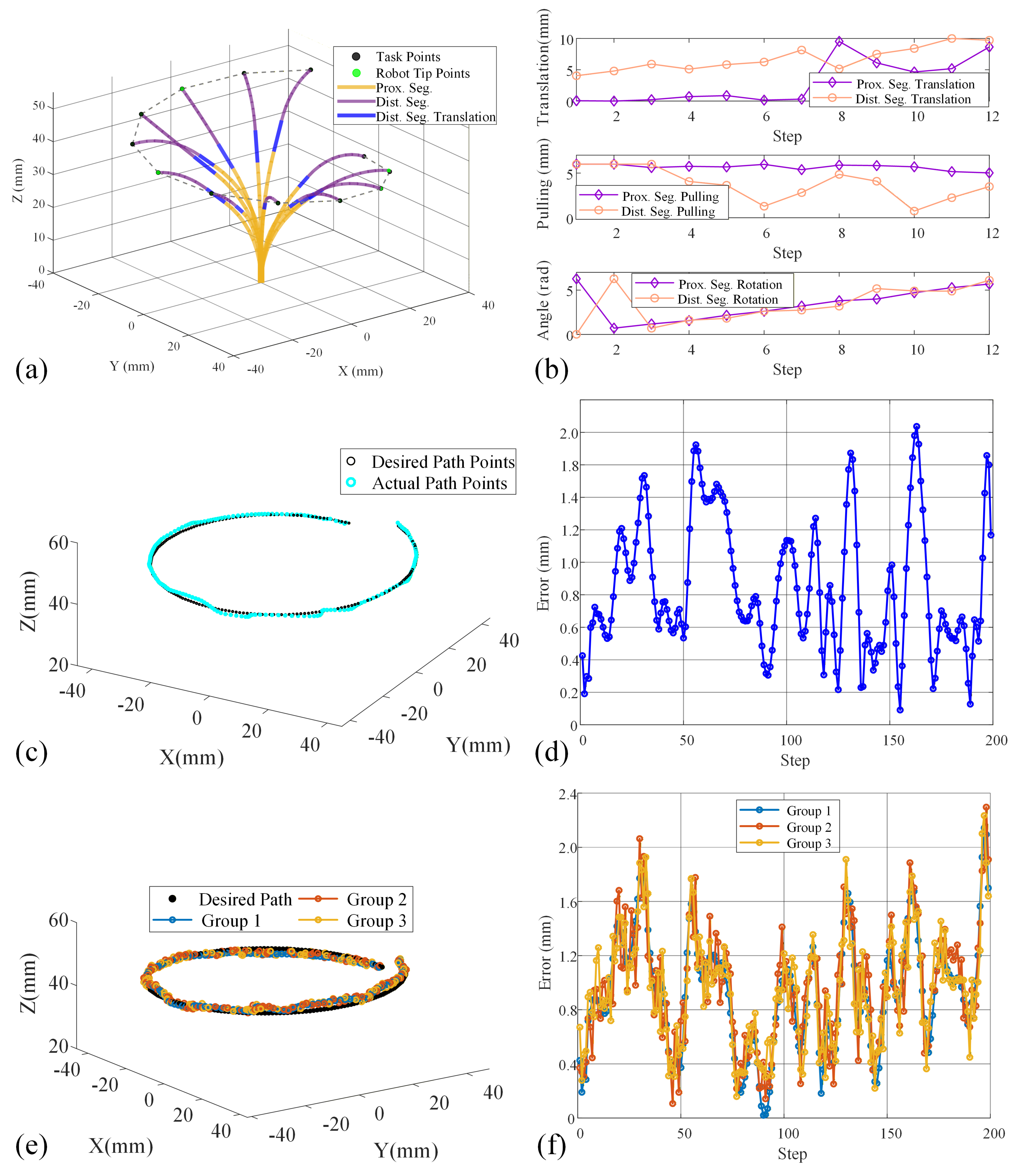}
	\caption{(a) Robot path-following task (simulation). (b) Variation of each DoF (simulation). (c) Robot end effector follows a circular path. (d) Tip position error in each step. (e) Additional three repeated experiments to validate the path-following accuracy. (f) Tip position errors of the three repeated experiments. }
	\label{fig:ikvalidation}
\end{figure}

\subsection{Manipulator Stiffness}
We then tested stiffness of the steerable section. Figure \ref{fig:stiffnessval} (a) shows the experiment setup. The dual-segment robot arm was fixed on the sensing system too, which was also calibrated to obtain the position of the tip and the midpoint. Both the two segments were actuated by setting a pulling distance of 3mm at the inner tube, and their bending angles reached around 45$^\circ$. Once a load is hang at the tip of the distal segment or the proximal segment, a deviation between the with-load and load-free conditions occurs. Intuitively, higher stiffness leads to smaller deviation.

To fully investigate the stiffness in the whole task space, upward, downward and leftward bending were all considered. As shown in Figure \ref{fig:stiffnessval} (b), the deviation at the tip points is related to the stiffness. Figure \ref{fig:stiffnessval} (c) shows the snapshots of the robot arm with hanging a 50g load and load-free scenarios. While we hang the load at the proximal segment's tip, the deviations of the upward, downward and left bending settings were 3.66mm, 3.31mm and 5.23mm. For brevity, only the snapshots of the robot carrying a load of 50g were shown in Figure \ref{fig:stiffnessval} (c). With the increase of load (varied from 50g to 300g), the deviation increased accordingly. Table \ref{tab:slits-offset} lists the deviation of each setting for the two segments. As a result, while the load reached 200g, 17.21mm of deviation was observed at the distal segment's tip. We found the maximum load-carrying capability of 300g at the downward bending setting. While the robot arm bent leftward, the load-carrying capability is lower. For PTE surgery, the maximum end effector's manipulation force of 1.5N could satisfy the clinical demands.
\begin{table*}
	\centering
	\small
	\renewcommand{\arraystretch}{1.2} 
	\caption{Deviation at the two segments' tip under different loads for tenon-mortise and square slits.}
	\label{tab:slits-offset}
	\begin{tabular}{ccccccccccccc}
		\toprule
		& \multicolumn{6}{c}{Load at Proximal Segment} & \multicolumn{6}{c}{Load at Distal Segment} \\
		\cmidrule(lr){2-7} \cmidrule(lr){8-13}
		\multirow{-2}{*}{\text{\begin{tabular}[c]{@{}c@{}}Tenon-mortise \\ slits\end{tabular}}} &
		\multicolumn{3}{c}{Distal Segment} &
		\multicolumn{3}{c}{Proximal Segment} &
		\multicolumn{3}{c}{Distal Segment} &
		\multicolumn{3}{c}{Proximal Segment} \\
		\midrule
		Load & \cellcolor[HTML]{F8CBAD}U & \cellcolor[HTML]{BDD7EE}D & \cellcolor[HTML]{C6E0B4}L & 
		\cellcolor[HTML]{F8CBAD}U & \cellcolor[HTML]{BDD7EE}D & \cellcolor[HTML]{C6E0B4}L & 
		\cellcolor[HTML]{F8CBAD}U & \cellcolor[HTML]{BDD7EE}D & \cellcolor[HTML]{C6E0B4}L & 
		\cellcolor[HTML]{F8CBAD}U & \cellcolor[HTML]{BDD7EE}D & \cellcolor[HTML]{C6E0B4}L \\
		50 g  & \cellcolor[HTML]{F8CBAD}6.41 & \cellcolor[HTML]{BDD7EE}3.70 & \cellcolor[HTML]{C6E0B4}9.91 & \cellcolor[HTML]{F8CBAD}3.66 & \cellcolor[HTML]{BDD7EE}3.31 & \cellcolor[HTML]{C6E0B4}5.23 & \cellcolor[HTML]{F8CBAD}4.91 & \cellcolor[HTML]{BDD7EE}5.67 & \cellcolor[HTML]{C6E0B4}5.74 & \cellcolor[HTML]{F8CBAD}4.93 & \cellcolor[HTML]{BDD7EE}4.22 & \cellcolor[HTML]{C6E0B4}3.98 \\
		100 g & \cellcolor[HTML]{F8CBAD}7.86 & \cellcolor[HTML]{BDD7EE}6.81 & \cellcolor[HTML]{C6E0B4}11.97 & \cellcolor[HTML]{F8CBAD}5.62 & \cellcolor[HTML]{BDD7EE}4.85 & \cellcolor[HTML]{C6E0B4}8.18 & \cellcolor[HTML]{F8CBAD}9.26 & \cellcolor[HTML]{BDD7EE}10.18 & \cellcolor[HTML]{C6E0B4}8.38 & \cellcolor[HTML]{F8CBAD}5.46 & \cellcolor[HTML]{BDD7EE}5.68 & \cellcolor[HTML]{C6E0B4}5.11 \\
		150 g & \cellcolor[HTML]{F8CBAD}11.21 & \cellcolor[HTML]{BDD7EE}9.16 & \cellcolor[HTML]{C6E0B4}14.68 & \cellcolor[HTML]{F8CBAD}9.62 & \cellcolor[HTML]{BDD7EE}6.15 & \cellcolor[HTML]{C6E0B4}8.35 & \cellcolor[HTML]{F8CBAD}14.20 & \cellcolor[HTML]{BDD7EE}13.32 & \cellcolor[HTML]{C6E0B4}18.06 & \cellcolor[HTML]{F8CBAD}9.35 & \cellcolor[HTML]{BDD7EE}8.16 & \cellcolor[HTML]{C6E0B4}7.96 \\
		200 g & \cellcolor[HTML]{F8CBAD}15.92 & \cellcolor[HTML]{BDD7EE}11.77 & \cellcolor[HTML]{C6E0B4} & \cellcolor[HTML]{F8CBAD}9.79 & \cellcolor[HTML]{BDD7EE}8.53 & \cellcolor[HTML]{C6E0B4} & \cellcolor[HTML]{F8CBAD} & \cellcolor[HTML]{BDD7EE}17.21 & \cellcolor[HTML]{C6E0B4} & \cellcolor[HTML]{F8CBAD} & \cellcolor[HTML]{BDD7EE}10.78 & \cellcolor[HTML]{C6E0B4} \\
		300 g & \cellcolor[HTML]{F8CBAD} & \cellcolor[HTML]{BDD7EE}14.51 & \cellcolor[HTML]{C6E0B4} & \cellcolor[HTML]{F8CBAD} & \cellcolor[HTML]{BDD7EE}10.96 & \cellcolor[HTML]{C6E0B4} & \cellcolor[HTML]{F8CBAD} & \cellcolor[HTML]{BDD7EE} & \cellcolor[HTML]{C6E0B4} & \cellcolor[HTML]{F8CBAD} & \cellcolor[HTML]{BDD7EE} & \cellcolor[HTML]{C6E0B4} \\
		\midrule
		& \multicolumn{6}{c}{Load at Proximal Segment} & \multicolumn{6}{c}{Load at Distal Segment} \\
		\cmidrule(lr){2-7} \cmidrule(lr){8-13}
		\multirow{-2}{*}{\text{\begin{tabular}[c]{@{}c@{}}Square\\  slits\end{tabular}}}  &
		\multicolumn{3}{c}{Distal Segment} &
		\multicolumn{3}{c}{Proximal Segment} &
		\multicolumn{3}{c}{Distal Segment} &
		\multicolumn{3}{c}{Proximal Segment} \\
		\midrule
		Load & \cellcolor[HTML]{F8CBAD}U & \cellcolor[HTML]{BDD7EE}D & \cellcolor[HTML]{C6E0B4}L & 
		\cellcolor[HTML]{F8CBAD}U & \cellcolor[HTML]{BDD7EE}D & \cellcolor[HTML]{C6E0B4}L & 
		\cellcolor[HTML]{F8CBAD}U & \cellcolor[HTML]{BDD7EE}D & \cellcolor[HTML]{C6E0B4}L & 
		\cellcolor[HTML]{F8CBAD}U & \cellcolor[HTML]{BDD7EE}D & \cellcolor[HTML]{C6E0B4}L \\
		50 g  & \cellcolor[HTML]{F8CBAD}18.92 & \cellcolor[HTML]{BDD7EE}17.09 & \cellcolor[HTML]{C6E0B4}17.60 & \cellcolor[HTML]{F8CBAD}8.72 & \cellcolor[HTML]{BDD7EE}7.41 & \cellcolor[HTML]{C6E0B4}7.55 & \cellcolor[HTML]{F8CBAD}37.11 & \cellcolor[HTML]{BDD7EE}27.31 & \cellcolor[HTML]{C6E0B4}31.39 & \cellcolor[HTML]{F8CBAD}14.72 & \cellcolor[HTML]{BDD7EE}8.17 & \cellcolor[HTML]{C6E0B4}9.65 \\
		100 g & \cellcolor[HTML]{F8CBAD}28.55 & \cellcolor[HTML]{BDD7EE}23.14 & \cellcolor[HTML]{C6E0B4}24.14 & \cellcolor[HTML]{F8CBAD}14.35 & \cellcolor[HTML]{BDD7EE}12.16 & \cellcolor[HTML]{C6E0B4}11.09 & \cellcolor[HTML]{F8CBAD} & \cellcolor[HTML]{BDD7EE}39.01 & \cellcolor[HTML]{C6E0B4} & \cellcolor[HTML]{F8CBAD} & \cellcolor[HTML]{BDD7EE}15.17 & \cellcolor[HTML]{C6E0B4} \\
		150 g & \cellcolor[HTML]{F8CBAD}38.68 & \cellcolor[HTML]{BDD7EE}32.82 & \cellcolor[HTML]{C6E0B4}28.13 & \cellcolor[HTML]{F8CBAD}19.86 & \cellcolor[HTML]{BDD7EE}15.29 & \cellcolor[HTML]{C6E0B4}15.71 & \cellcolor[HTML]{F8CBAD} & \cellcolor[HTML]{BDD7EE} & \cellcolor[HTML]{C6E0B4} & \cellcolor[HTML]{F8CBAD} & \cellcolor[HTML]{BDD7EE} & \cellcolor[HTML]{C6E0B4} \\
		200 g & \cellcolor[HTML]{F8CBAD} & \cellcolor[HTML]{BDD7EE}39.68 & \cellcolor[HTML]{C6E0B4} & \cellcolor[HTML]{F8CBAD} & \cellcolor[HTML]{BDD7EE}21.29 & \cellcolor[HTML]{C6E0B4} & \cellcolor[HTML]{F8CBAD} & \cellcolor[HTML]{BDD7EE} & \cellcolor[HTML]{C6E0B4} & \cellcolor[HTML]{F8CBAD} & \cellcolor[HTML]{BDD7EE} & \cellcolor[HTML]{C6E0B4} \\
		300 g & \cellcolor[HTML]{F8CBAD} & \cellcolor[HTML]{BDD7EE} & \cellcolor[HTML]{C6E0B4} & \cellcolor[HTML]{F8CBAD} & \cellcolor[HTML]{BDD7EE} & \cellcolor[HTML]{C6E0B4} & \cellcolor[HTML]{F8CBAD} & \cellcolor[HTML]{BDD7EE} & \cellcolor[HTML]{C6E0B4} & \cellcolor[HTML]{F8CBAD} & \cellcolor[HTML]{BDD7EE} & \cellcolor[HTML]{C6E0B4} \\
		\bottomrule
	\end{tabular}

	{\footnotesize \textbf{Note:} U = Upward, D = Downward, L = Leftward. Unit: mm. Blank means the robot arm failed to lift the load.}
\end{table*}
To further investigate the notable stiffness, we manufactured another batch of tubes whose slits are square shape. The slit width is identical with that the tenon-mortise tubes. We repeated the above experiments, and results show that the deviation was much larger, as listed in Table \ref{tab:slits-offset}. The square-slit robot arm failed to offset the load of 300g. While the load is at the distal segment's tip, it could just lift the load of 100g, which obviously demonstrates the higher stiffness of the tenon-mortise tubes. The process of hanging the load for all the settings is also recorded in the Supplementary Video.
\begin{figure}
	\centering
	\includegraphics[width=1\linewidth]{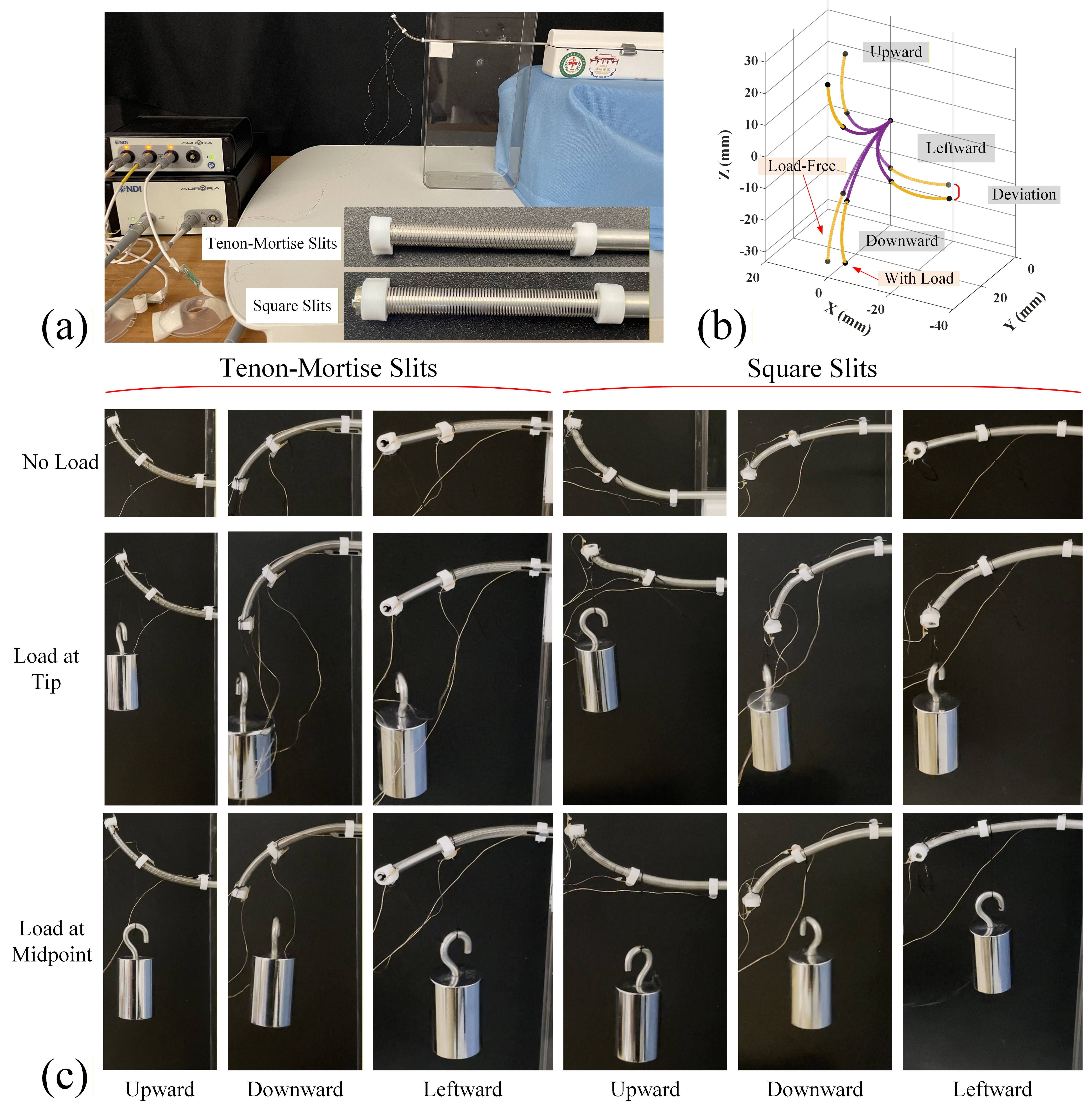}
	\caption{Stiffness validation. (a) Experiment setup. We hang a series of loads at the tip of the two segments respectively to investigate the tip deviation in with-load and load-free scenarios. Tubes with tenon-mortise slits and square slits were also used for comparison. (b) Illustration the three bending modes, i.e., upward, downward and horizontal. The deviation at the tip point shows the stiffness. (c) Snapshots of the with-load (50g) and load-free conditions.}
	\label{fig:stiffnessval}
\end{figure}

\begin{figure}
	\centering
	\includegraphics[width=1\linewidth]{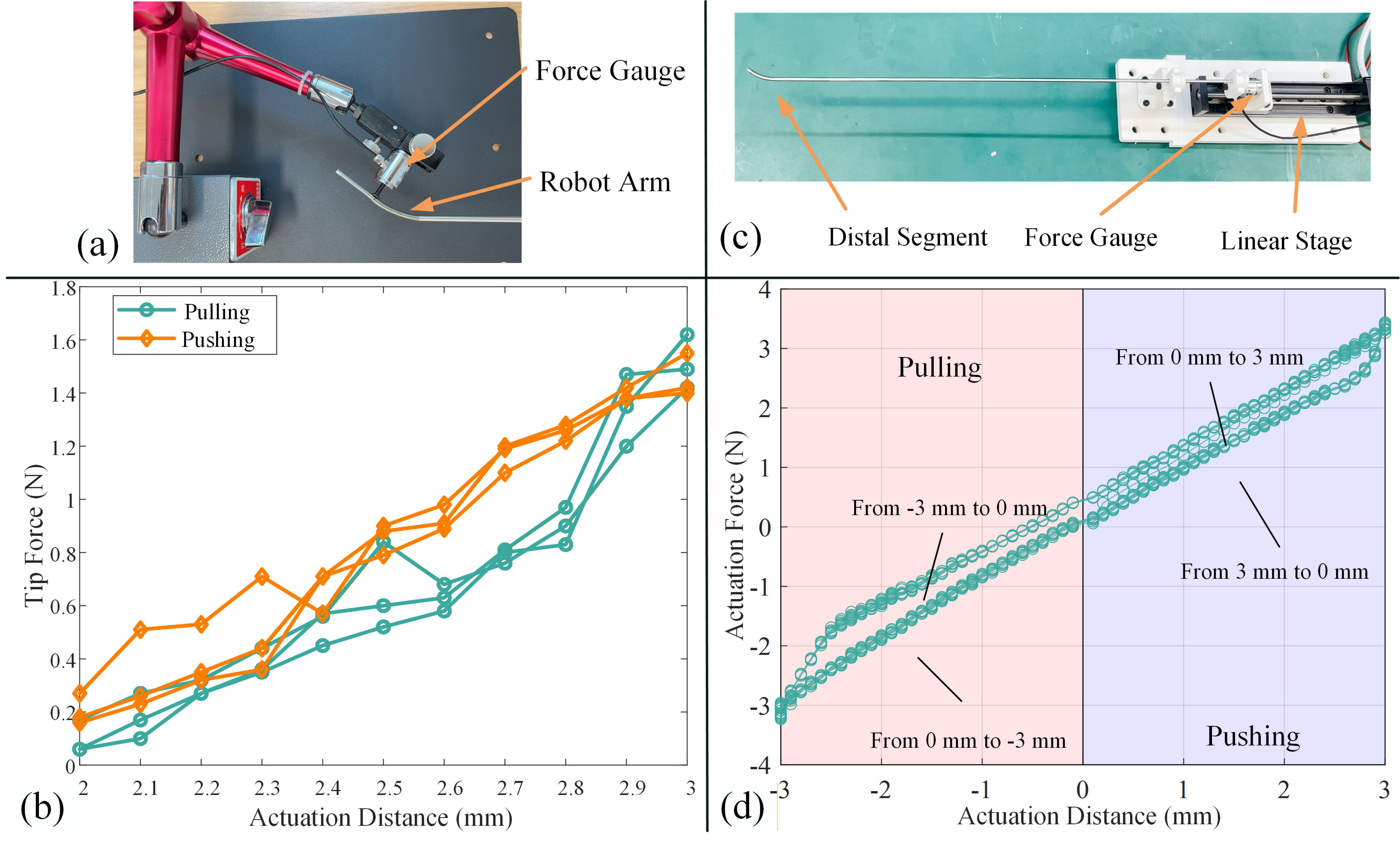}
	\caption{ (a) Experiment setup to investigate the tip force and push/pull distance. (b) Three sets of data about the actuation distance and tip force for both the proximal and the distal segments. (c) Experimental setup to measure the push/pull force acting on the inner tube. (d) Actuation force with different actuation distance states.}
	\label{fig:forcesen}
\end{figure}

The manipulation force at the tip and actuation were quantitatively measured. Figure \ref{fig:forcesen} (a) shows the experiment setting, where we focused on radial force at the tip. A force gauge (SBT641-19.6N, SIMBATUOCH, China) was fixed on the experimental platform, and the proximal segment was actuated to approach it. The pulling distance increased from 2mm to 3mm with a 0.1mm increment. The shape variation is blocked by the force gauge, so the sensor readings increased accordingly. Figure \mbox{\ref{fig:forcesen}} (b) shows the sensor readings with each actuation distance ranging from 2mm to 3mm. The tip force and actuation distance show a relatively linear relationship in multiple repeated experiments, indicating that the larger the bending angle, the higher the stiffness of the robot. The maximum tip acting force reached $1.7$N, which is sufficient for handling most tissues. Finally, we measured the actuation force (push force or pulling force on the inner tube in bending), and the experiment setup is shown in  Figure \mbox{\ref{fig:forcesen}} (c). The force gauge was connected with the inner tube, and the linear stage generated actuation distance. Only the distal segment was tested about the actuation force for it is directly used for manipulation. The inner tube of the distal segment was connected to the gauge and actuated within $\pm$ $3$mm range ($0.1$mm gap). The pulling or pushing force at the actuation sides was recorded after each instance. A 16-hour fatigue test (2000 cycles) confirmed system durability, with no mechanical breakage observed. Part of the testing process was recorded in the Supplementary Video.
Figure \mbox{\ref{fig:forcesen}} (d) shows the variation of the actuation force, demonstrating the repeatability. In surgery, the instrument component is designed for single use, and the surgery time only last for several hours. The high repeatability also demonstrates the feasibility and durability.
\subsection{Dexterity Test}
In terms of steerability and ergonomics, basic clinic potential was tested in a bifurcate pipe setting, with red water mimicking the residual blood in PA. As shown in Figure \ref{fig:dexterityval} (a), the handheld dissector was inserted from a vertical port. Then, a pump started to work, and the red water polluting camera view was aspired from the instrument channel, which was recorded in Supplementary Video. While facing the bifurcate intersection, the steerable section was manipulated to move downwards. This experiment demonstrates its accessibility in tortuous task space and that the instrument channel could be used for suctioning residual blood.
\begin{figure}
	\centering
	\includegraphics[width=1\linewidth]{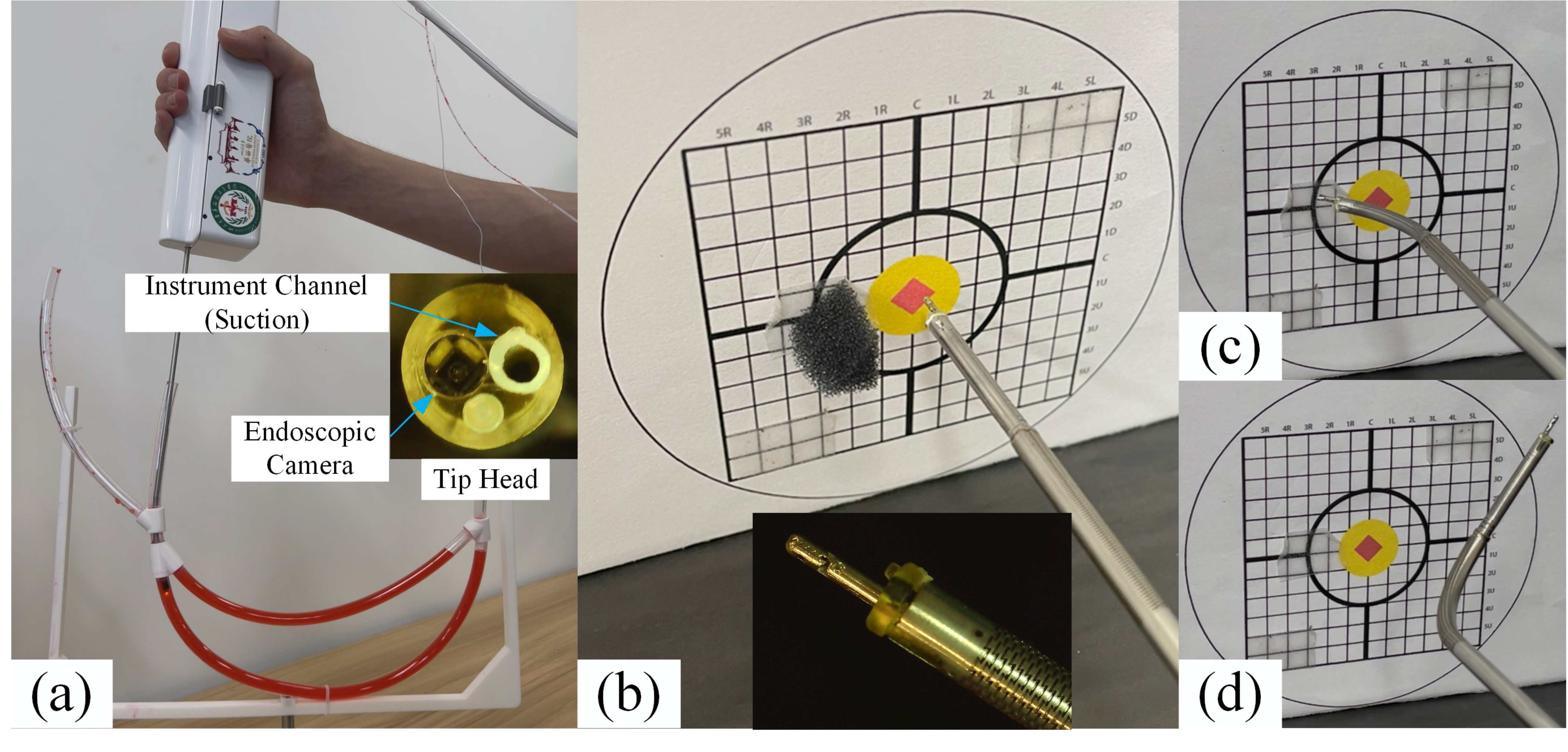}
	\caption{Dexterity test. (a) Blood suction simulation and steerablity test in bifurcate pipe. (b) Target manipulation test setting, and a biopsy gripper extended from the instrument channel. (c) Distal segment bends. (d) The two segments are both bending to position the tip tool.}
	\label{fig:dexterityval}
\end{figure}
Furthermore, the steerability and maneuverability were evaluated on an additional experimental setting, shown in Figure \ref{fig:dexterityval} (b). A slender biopsy forceps passes through the instrument channel, and a block of foam simulates the to-be-stripped tissue.  Before manipulation, the distal segment was individually commanded to explore the circled area while the proximal segment kept straight, demonstrating its reachability, as shown in Figure \ref{fig:dexterityval} (c). However, the dexterity is limited, so the two segments bent concurrently, shown in Figure \ref{fig:dexterityval} (d). It not only covered the whole task space, but also presented multiple shapes and notable dexterity. Opening and closing of biopsy forceps were also controlled by operator's right hand.

In addition to handheld application, the dissector could also be mounted on a rigid robot arm (xMate ER7 Pro, ROKAE, China) to realize full robotization, as shown in Figure \ref{fig:pteinstru} (b). This setting could be applied in other similar surgeries, such as with-radiation surgeries where surgeon could teleoperate the slender robot arm under the guidance of X-ray images. The overall setting and manipulation are shown in the Supplementary Video. In an operation room, the rigid robot arm brings the whole dissector beside patient and only feed along the $Z$ axis of the dissector's base frame. Consequently, the dissector could enter the surgical site, and the whole process could be operated in a master-slave manner by a surgeon seated beside the  console.
\subsection{Explore Tortuous and Confined Space}
We then validated its capability in accessing tortuous confined space, as shown in Figure \ref{fig:confinedspaceexplore}. First, we used a transparent bronchus model to simulate accessing depth of PA, and it should be noted that the main branch of the plastic bronchus model is similar to that of PA in size, and the deep branches of PA is much thinner. The reason why we used this 3D model is availability and its transparency. An operator held the dissector to explore the tortuous shape of the transparent model, starting from the main branch (Figure \ref{fig:confinedspaceexplore} (a)). While meeting the first bifurcation, the operator steered the distal segment to bend and turn left to the orient the second-class branch  (Figure \ref{fig:confinedspaceexplore} (b)), and the endoscopic camera provided clear view of the scene. It further moved forward and the bending angle of the two segments increased, so the endoscopic camera could see the scene of the third-class branch. This exploration also demonstrates that the dual-segment device has notable accessibility. The whole process of entering and exiting only took 10 seconds.

\begin{figure}
	\centering
	\includegraphics[width=1\linewidth]{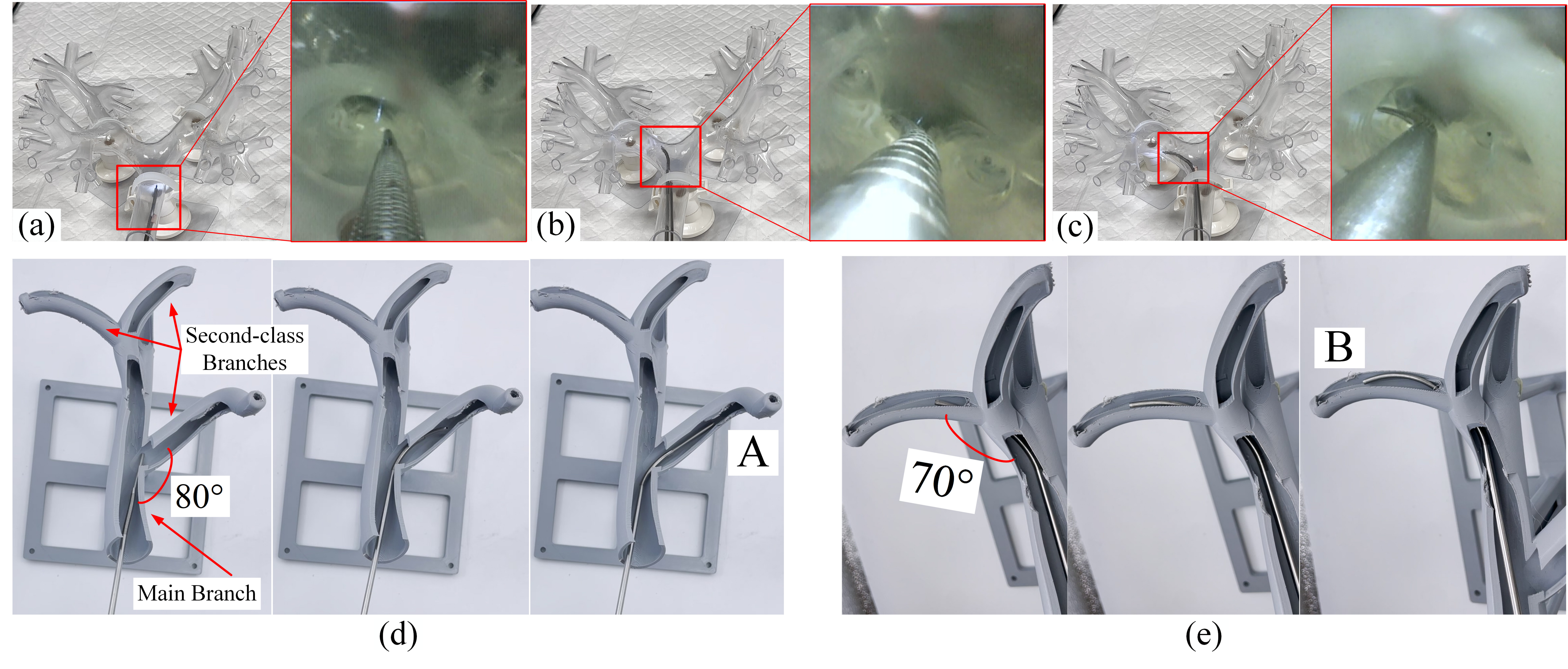}
	\caption{Explore confined unstructured task space. (a) Use a transparent bronchus model to test the capability of exploring tortuous confined task space. (b) Turn left from the main branch to reach the second-class branch. (c) Continuously advance to the deep branch and the endoscopic camera saw the scene of the third-class branch. (d) A 3D-printed tortuous vessel model was employed to test the dexterity, and the second class of branch has a diameter of 9mm.  (e) Exploring another second class branch with a diameter of 6mm.}
	\label{fig:confinedspaceexplore}
\end{figure}

To mimic the shape of PA that is tortuously distributed in human lung, we prepared a 3D-printed PA model to simulate PTE. It has a main branch and three second-class branches, as shown in Figure \ref{fig:confinedspaceexplore} (d). For visualizing the dissector's shape, the model was designed semi-closed. Also, the operator held the dissector by his right hand and held the control unit using the left hand. The steerable section entered from the main branch and turned right to reach target A (shown in Figure \ref{fig:confinedspaceexplore} (d)), where the included angle between the second-class branch and the main branch was around $80^\circ$. As a result, by controlling the bending angle of the two segments, the steerable section easily reached the target point, and the whole process only took 12s. Then, we tested reaching goal B, as shown in Figure \ref{fig:confinedspaceexplore} (e). The included angle was around $70^\circ$, but the length is around 6cm. Finally, the tip of the dissector also reached the desired target easily, which took 21s. In PTE, PA is soft and the dissector is likely to enter the third-class arteries. In the following, we will conduct ex vivo experiments to further test the dissector's clinical potential. 
\subsection{Ex Vivo Trial}
\begin{figure}
	\centering
	\includegraphics[width=1\linewidth]{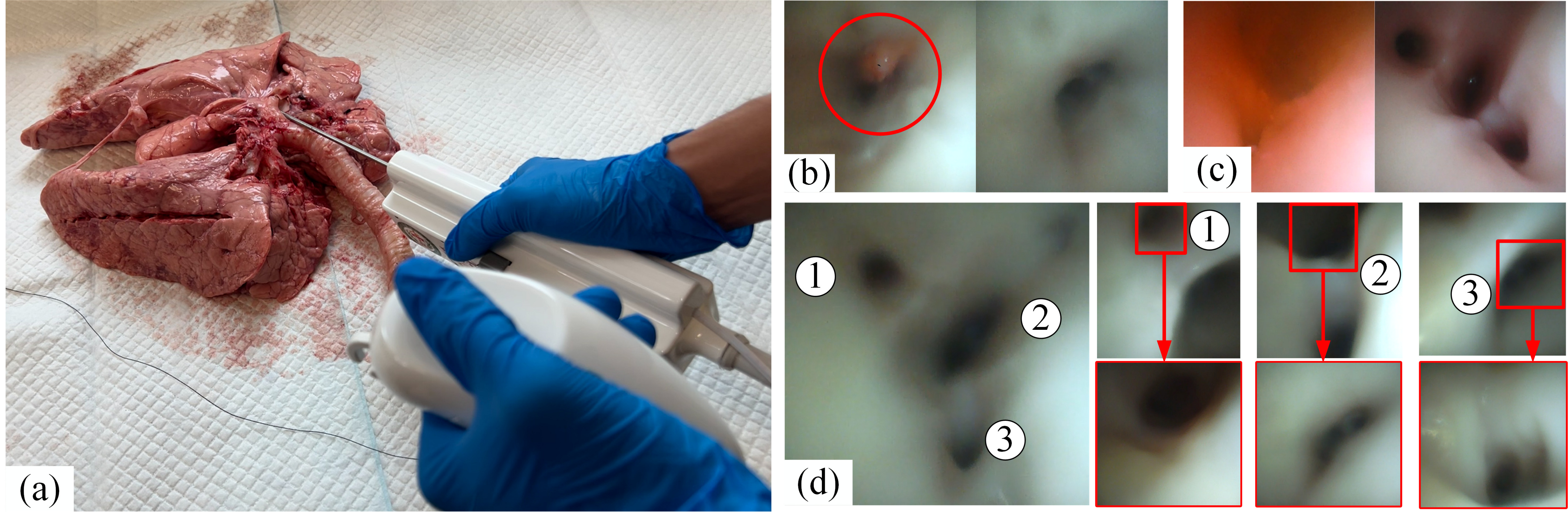}
	\caption{(a) Ex vivo trial experiment setup.  (b) Surgical site before and after stripping tumor. (c) Residual blood before and after suction. (d) Path exploration. In the endoscopic view, three thin PA branches were explored by the dissector.}
	\label{fig:ExVivoTrial}
\end{figure}
To further validate the effectiveness of the dissector in PTE, a porcine lung was leveraged to conduct ex vivo trial, which simulates the actual surgery. Three aspects were considered in this experiment, i.e., tumor removal simulation, residual blood suction, and tortuous path exploration in PA. A fresh porcine lung was placed on a table and an operator (the first author) used scissors to establish a small opening at the main branch of PA. His left hand held the control unit and the right hand held the dissector, as shown in Figure \ref{fig:ExVivoTrial} (a). The whole process was supervised by a professional surgeon (the 5th author). Next, we placed a small piece of porcine tissue at the second-class branch to simulate the tumor in PA. With the assistance of endoscopic view, the steerable section was easily controlled to enter the intersection between the first branch and the second. Soon, the tissue was observed, and the corresponding endoscopic view is shown in Figure \ref{fig:ExVivoTrial} (b). After identifying the tissue, the distal segment slowly approached it. With delicate manipulation, the tissue was stripped from PA, and the whole process only last around 30s. Before and after removal the tissue of endoscopic views are shown in Figure \ref{fig:ExVivoTrial} (b).

Then, a small amount of red water was filled into PA to simulate residual blood. Therefore, the dissector could suction the residual blood. A suction pump connecting the instrument channel always worked for generating vacuum pressure. Figure \ref{fig:ExVivoTrial} (c) shows the endoscopic views of before and after suction. It also provides clear view for surgeon with the irrigated saline.

At last, the curve path from the first branch to the fourth branch was explored for further testing. Figure \ref{fig:ExVivoTrial} (d) shows the view of intersection between the first branch and the second, from which we can see three holes (labeled as \Circled{1}, \Circled{2}, and \Circled{3}) to access to the second branch. Operator controlled the distal segment to steer its direction from orienting \Circled{2} to orienting \Circled{1}. Then, the translation DoF of the distal segment worked to enter hole \Circled{1}. Similarly, the dissector moved from \Circled{1} to \Circled{2}, and from PA \Circled{2} to  \Circled{3}, respectively. The whole process was recorded in the Supplementary Video. It demonstrates that steerability at the tip is of great significance in clinic application. The curved steerable section enables accessing into a distal thin PA branch easily, which is more effective than using conventional rigid dissectors. 
\begin{figure}
	\centering
	\includegraphics[width=1\linewidth]{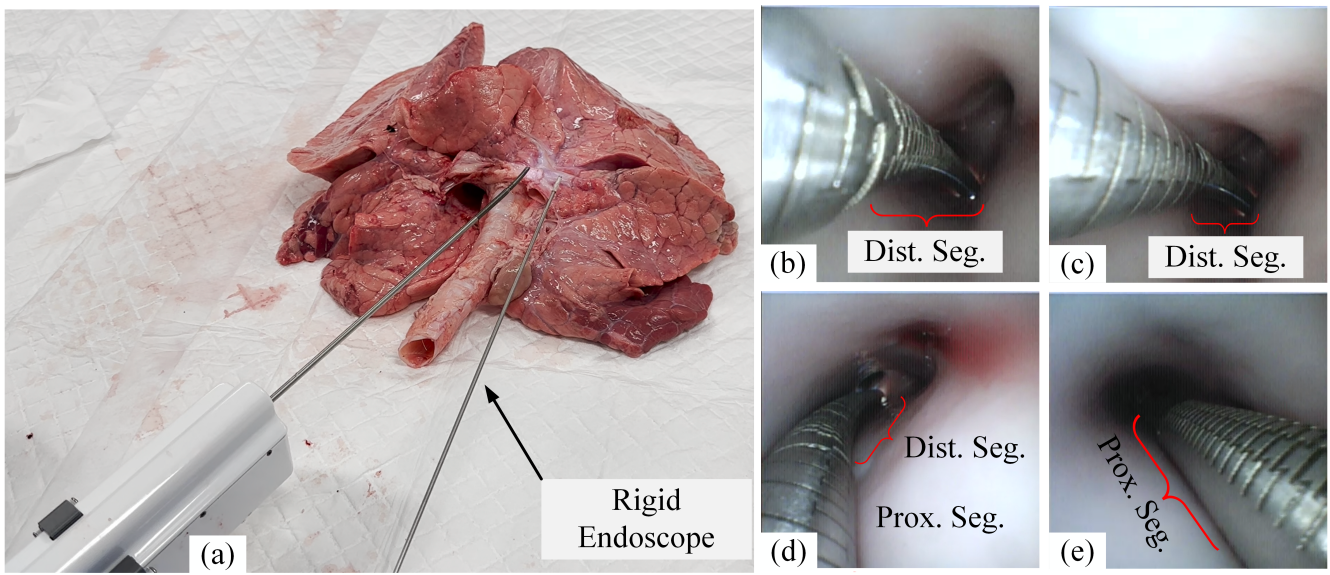}
	\caption{Dissector's shape in surgery simulation.  (a) A rigid endoscope is also inserted into PA to see the dissector's shape. (b)-(e) Shapes of the dissector in four different PA branches. }
	\label{fig:ExvivoTrialTwoEndo}
\end{figure}

To collect the shape of the whole steerable section, a rigid endoscope was employed as a follower to track the motion of the dissector, enabling a third-person perspective view. It should be noted that the straight endoscope is not needed in actual surgery. Herein, we just hope to collect the shape of the dissector. During the whole process, we tried to control the rigid endoscope to stay beside the proximal segment to obtain better view. Initially, the distal segment actively bent within the secondary PA cavity, as shown in Figure \ref{fig:ExvivoTrialTwoEndo} (b). It continued to enter the deeper branches (see Figure \ref{fig:ExvivoTrialTwoEndo} (c)). Before entering the third-class PA, the bending angles of both the proximal and distal segments were observed in Figure \ref{fig:ExvivoTrialTwoEndo} (d). However, due to the thin third-class PA and the rigid endoscope's inability to actively bend, the bending shape of the distal segment could not be recorded effectively at this area, as shown in Figure \ref{fig:ExvivoTrialTwoEndo} (e). Actually, the two segments were commanded to bend to enter the depth of PA. 

\subsection{Discussion}\label{Discuss}
The experimental results demonstrate the dissector’s capability of exploring complex and thin branches of pulmonary artery (PA), with kinematic validation showing a 2mm accuracy in open-loop control scheme and comparable stiffness to withstand up to 300g load. Ex vivo trials confirmed its clinical potential in thrombus removal, suction, and serpentine PA exploration, achieving tissue dissection and effective path navigation within thin PA branches. The integrated suction channel and ergonomic control unit further validated the feasibility. We have miniaturized the size of the dissector to 3.5mm in diameter and integrated three instrument channels inside it, i.e., endoscopic camera's signal wire, saline irrigation passage and a channel for passing the gripper. The notable advantages are compactness in manipulation and dexterity in accessing tortuous surgical sites. The slender dual-segment robotic device has six DoFs, providing notable dexterity in confined task space. For PTE surgery, this dexterous dissector enables entering the deep branches of PA effectively and the endoscopic manipulation benefits safe and complete clots removal. Experiments on 3D-printed models and ex vivo porcine lung demonstrate that the time cost is around 1 minutes (starting from entering the main branch of PA and ending still reached the third-class branch). As an estimation, endarterectomy procedure may take 5 minutes since the endoscopic view will provide guidance and feedback for the surgeon, while standard PTE roughly takes 20 minutes for one turn of clots removal \cite{madani2016surgical, jamieson1998pulmonary}. On the other hand, this robotized dissector could follow the tortuous path of PA owing to the dexterity, which almost causes no trauma while manipulating the clots in thin branches. It also should be noted that other procedures are identical with standard PTE and our work only focuses on enhancing the completeness and effectiveness of clot removal procedure. The steerable section is also able to accommodate other similar instruments, like a an electrode to cauterize tissues and to stop bleeding. Therefore, this design may be potential for neurosurgery, orthopedics surgery and trans-nasals surgeries. 

Although the success of ex vivo trial paves a way for subsequent in vivo trial, several limitations remain to be addressed: 1) the endoscopic camera’s resolution and field of view require enhancement for better visualization in blood-filled environments; 2) ex vivo trials on porcine lungs may not fully replicate human PA's complexity and size; and 3) the absence of haptic feedback limits force-sensitive maneuvers in fragile tissues, and the learning curve for surgeons will be longer than the conventional rigid straight dissectors. In the future, we will focus on integrating high-definition imaging, force-sensing capabilities, and elasticity materials (e.g., NiTi alloy) to improve durability and safety. These advancements aim to bridge the gap between ex vivo validation and clinic trials.
\section{Conclusion}\label{Conclusion}
In this work, we have proposed a novel steerable multi-function dissector for endoscopic pulmonary thromboendarterectomy, providing endoscopic view and high level of steerability in tortuous pulmonary artery. This significantly enhanced surgery outcome, including removing clots in distal thin branch of PA, reducing surgery time and decreasing complexity. The proposed optimization-based inverse kinematics model is effective for solving the actuation configuration, and considered the condition that the distal segment nests into the proximal. To the best of our knowledge, there is no similar work on developing robotized instruments for PTE surgery. Based on concentric push/pull tubular robot, we proposed tenon-mortise slits along the steerable section to enhance stiffness, and finally it could offset load of 300 g.  Ex vivo experiments in a porcine lung demonstrate that the dissector is able to enter the fourth-class branch and could remove the thromb and intima inside PA. We believe that the steerable dissector will pave a novel way for design surgical instruments applied in minimally invasive surgeries. 

\section*{Acknowledgments} 
This work was partially supported by Sichuan Science and Technology Program (Grant number: 2023YFH0093).

\section*{Author Disclosure Statement}
No competing interests exist.

\bibliographystyle{ieeetr}
\bibliography{references.bib}
}
\end{multicols}

\end{document}